\pdfoutput=1

\documentclass[11pt]{article}

\usepackage[final]{acl}

\usepackage{times}
\usepackage{latexsym}

\usepackage[T1]{fontenc}

\usepackage[utf8]{inputenc}

\usepackage{microtype}

\usepackage{inconsolata}

\usepackage{tikz,graphics,color,fullpage,float,epsf,caption,subcaption,booktabs,amsmath,multirow, array}

\title{Preserving Pre-trained Representation Space:\\ On Effectiveness of Prefix-tuning for Large Multi-modal Models}

\author{
    Donghoon Kim,
    Gusang Lee,
    Kyuhong Shim, and 
    Byonghyo Shim \\
    {\normalsize Dept. of Electrical and Computer Engineering, Seoul National University} \\
    {\texttt{\small \{dhkim,gslee,khshim,bshim\}@islab.snu.ac.kr}} \\
}

\begin{document}
\maketitle
\pagestyle{empty}
\begin{abstract}
Recently, we have observed that Large Multi-modal Models (LMMs) are revolutionizing the way machines interact with the world, unlocking new possibilities across various multi-modal applications.
To adapt LMMs for downstream tasks, parameter-efficient fine-tuning (PEFT) which only trains additional prefix tokens or modules, has gained popularity.
Nevertheless, there has been little analysis of how PEFT works in LMMs.
In this paper, we delve into the strengths and weaknesses of each tuning strategy, shifting the focus from the efficiency typically associated with these approaches.
We first discover that model parameter tuning methods such as LoRA and Adapters distort the feature representation space learned during pre-training and limit the full utilization of pre-trained knowledge.
We also demonstrate that prefix-tuning excels at preserving the representation space, despite its lower performance on downstream tasks.
These findings suggest a simple two-step PEFT strategy called \textbf{Prefix-Tuned PEFT (PT-PEFT)}, which successively performs prefix-tuning and then PEFT (i.e., Adapter, LoRA), combines the benefits of both.
Experimental results show that PT-PEFT not only improves performance in image captioning and visual question answering compared to vanilla PEFT methods but also helps preserve the representation space of the four pre-trained models.


\end{abstract}
\section{Introduction}
\label{sec:intro}
Understanding the visual scene and expressing it with a natural language are two distinct tasks yet the human brain can comprehensively handle both without difficulty.
Large multi-modal models (LMMs) mimic such capability by training a deep neural network (DNN) such that it learns semantically meaningful connections between vision and language from a large number of image-text pairs~\citep{li2020oscar,zhang2021vinvl,wang2021simvlm,radford2021clip}.
Recently, LMMs have been widely used due to their broad range of applications, including chatbot, robot control, and video generation~\citep{ouyang2022chatgpt,brohan2023robot_vl,ramesh2022dalle2}.
\begin{figure}[t!]
\centering
\includegraphics[width=0.9\linewidth]{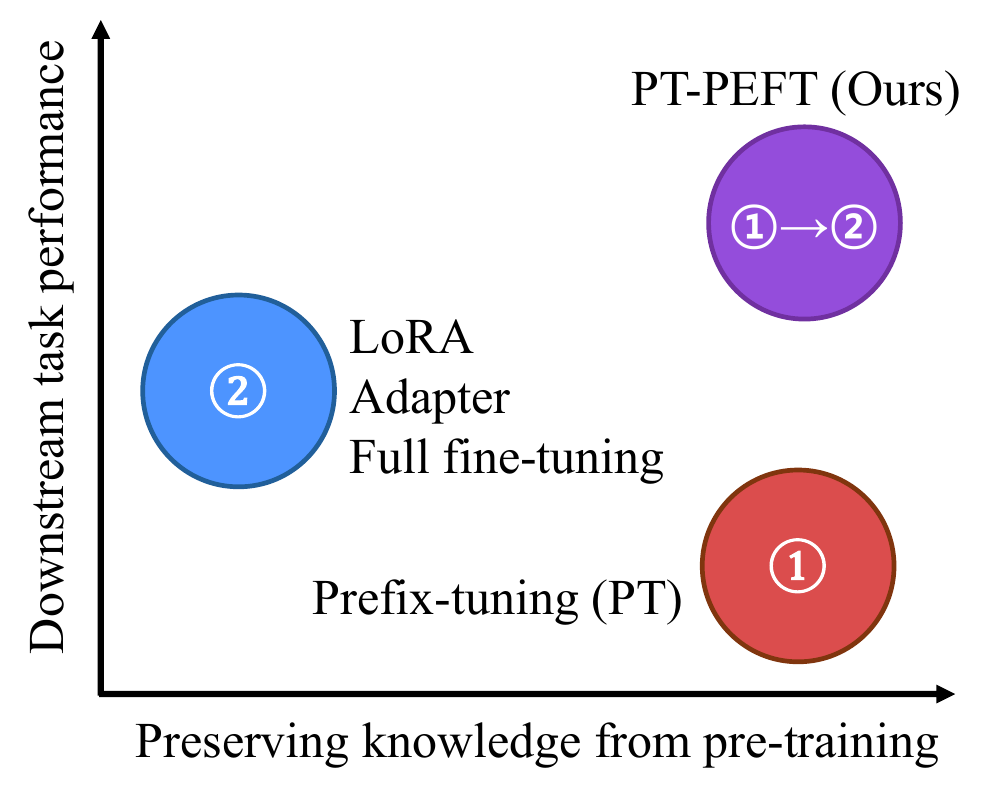}
\caption{Advantages of the proposed PT-PEFT, which performs 1) prefix-tuning and 2) fine-tuning (i.e., parameter-efficient or full fine-tuning) sequentially.}
\label{fig:pt-peft}
\end{figure}

In the \textit{pre-training}, LMMs are trained to predict the masked words or next words from the image-text pair~\citep{li2023blip2,alayrac2022flamingo,wang2022ofa}.
In the second step called \textit{fine-tuning}, the pre-trained LMMs are tailored to the specific downstream task.
It has been shown that fine-tuning provides superior performance in various downstream tasks such as image captioning (IC), visual question answering (VQA), and image-text retrieval~\citep{li2023blip2,wang2022ofa,wang2021simvlm,zhang2021vinvl}.
However, fine-tuned models often suffer from the loss of generalization capability obtained from the pre-training \citep{sun2015surveyDA,brown2020gpt3}.
Since the task-specific dataset is far smaller than the pre-training unlabeled dataset, the pre-trained model can be easily overfitted to the small-sized downstream task dataset, leading to degraded performance~\citep{kumar2022finetune_distort}.
Various approaches have been suggested over the years to address the problem.
In prompt-based approaches, manually designed prompts or trainable continuous embedding vectors are integrated into the input data to adapt the model for downstream tasks~\citep{li2021prefix,liu2021p-tuningv2,tam2022prompt,lester2021power_prompt}.
In knowledge distillation-based fine-tuning approaches, the model minimizes the distance between the distribution of the pre-trained and fine-tuned models~\citep{xu2020bert_self_distil,sanh2019distilbert,boschini2022transfer}.
The common wisdom behind these approaches is to minimize the modification of the pre-trained model parameters while maintaining performance on downstream tasks.
\begin{figure}[t!]
    \centering
    \includegraphics[width=1.0\linewidth]{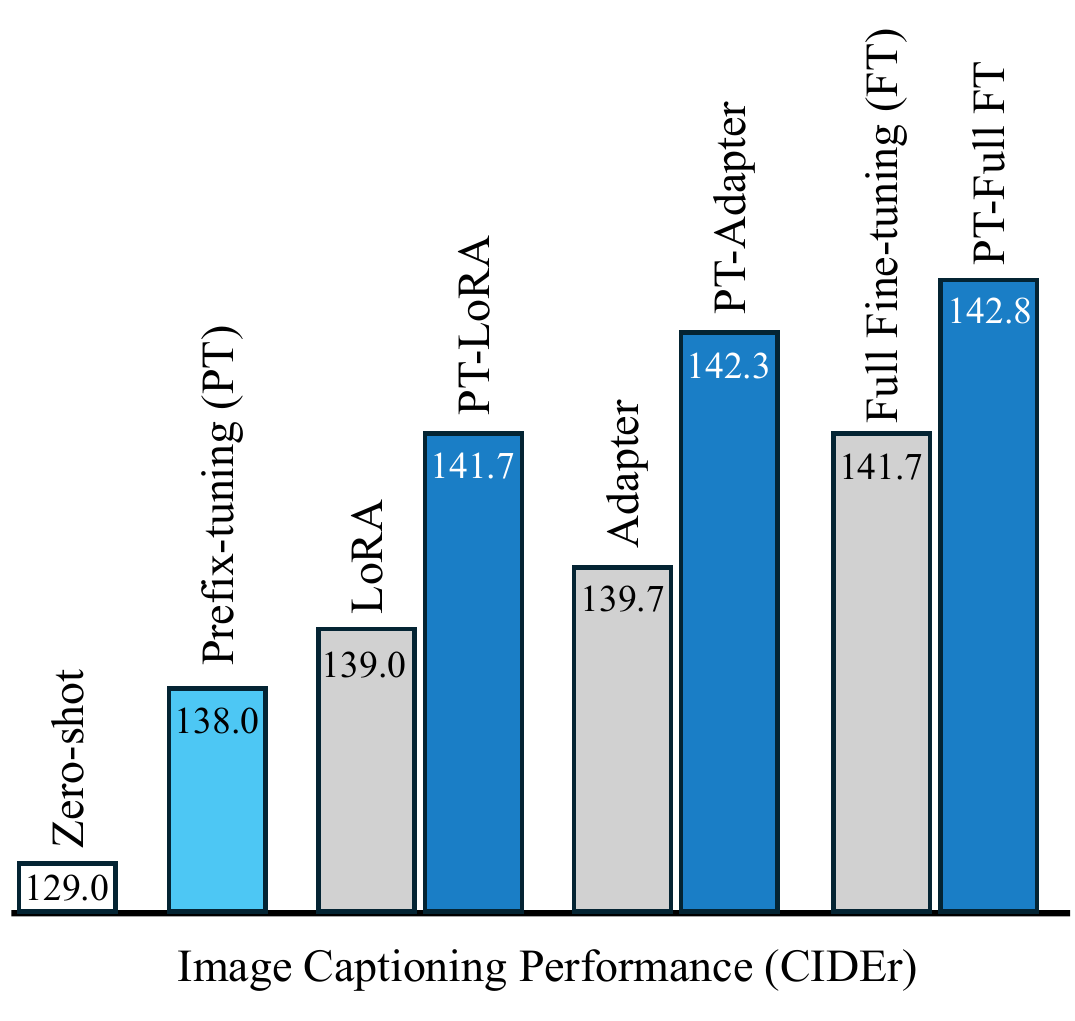}
    \caption{
        Performance of different task adaptation methods on COCO image captioning dataset. The proposed method (PT-) consistently improves performance when combined with other methods.}
    \label{fig:imt_overivew}
\vspace{-0.5cm}
\end{figure}

One drawback of the full model fine-tuning is the huge computational burden caused by the model parameters update.
In an effort to reduce the huge training cost, various parameter-efficient fine-tuning (PEFT) techniques have been proposed~\citep{li2021prefix,houlsby2019adapter,hu2021lora,he2021adaptereffectiveness}.
In these approaches, only a small set of additional modules (e.g., prefix, Adapter, LoRA) is trained instead of relying on full fine-tuning.
These approaches are especially beneficial for training the large pre-trained model like GPT~\citep{gpt-3}, T5~\citep{t5-lm}, and Llama~\citep{touvron2023llama2}.

Training efficiency is a well-known advantage of prefix-tuning. 
Unlike other PEFT methods, prefix-tuning does not modify the model's parameters, leaving the representation space unchanged.
To investigate the changes in the representation space, we analyze the feature representation matrices using singular value decomposition (SVD).
Notably, we observe that the representation space of a fine-tuned model (in IC and VQA) utilizes only a limited set of effective basis vectors (60\% of those in the pre-trained model) to express the output.
Clearly, this limits the model's ability to fully enjoy the benefits obtained from pre-training (see Figure~\ref{fig:svd_ranks}).
In contrast, we discover that all the basis vectors are utilized in the prefix-tuned model, implying that the prefix-tuning effectively preserves the inherited representation space from the pre-training.

While the prefix-tuning is effective in preserving pre-trained knowledge, the efficacy of this approach is somewhat questionable since the reported evaluation results are not conclusive.
Some studies claim that the prefix-tuning performs comparable to the model parameter-tuning (e.g., full fine-tuning, LoRA, Adapter), while others argue that the prefix-tuning struggles in the training of relatively small-sized language models~\citep{liu2021p-tuningv2,tam2022prompt}.
\begin{figure*}[t!]
    \centering
    \includegraphics[width=1.0\textwidth]{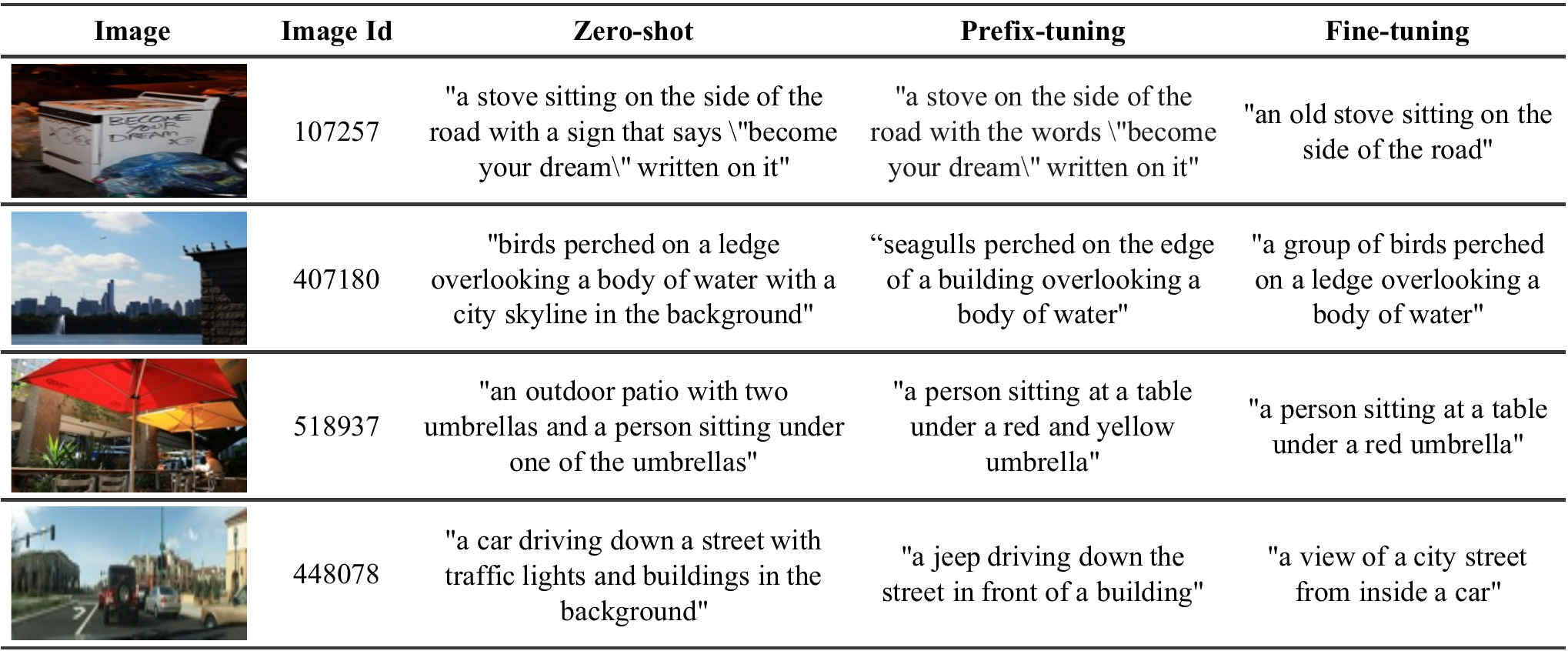}
    \caption{Qualitative image captioning results of zero-shot learning, prefix-tuned, and fine-tuned models. Although fine-tuning provides accurate answers, its results often ignore visual details compared to the other two.}
    \vspace{-0.1cm}
    \label{fig:zero}
\end{figure*}

An aim of this paper is to propose a simple yet effective tuning strategy to combine the merits of two seemingly distinct approaches.
The proposed method, henceforth referred to as Prefix-Tuned PEFT (PT-PEFT), performs the prefix-tuning and the model parameter-tuning \textit{sequentially}.
The key feature of PT-PEFT is to preserve the pre-trained feature space through the prefix-tuning and then refine the model parameters using the PEFT method.
Intuitively, this approach resembles a language model learning a new task using prompt sentences such as "I will provide example sentences describing the given pictures in the news article style. So, please generate the caption for the given images with such style."
By providing a context suitable for the new task, the model's adaptability is enhanced, allowing for faster convergence and minimal changes to the weights of the pre-trained model.

In our experiments, we show that applying the prefix-tuning before LoRA, Adapter, and even full fine-tuning consistently improves the task performance for all datasets and various pre-trained LMMs including BLIP~\citep{li2022blip}, BLIP-2~\citep{li2023blip2}, OFA~\citep{wang2022ofa} and VINVL~\citep{zhang2021vinvl}.
We also compare the simultaneous tuning of prefix and model parameters and show that the proposed sequential strategy is indeed important for maximizing performance and preserving the representation space.

\vspace{0.2cm}
Our contributions are as follows:
\begin{itemize}

    \item We show the correlation between the representation space and performance through rank-based analysis. We qualitatively and quantitatively illustrate the adverse effects of representation space collapse in task performance.
    
    \item We reveal that the prefix-tuning differs significantly from model parameter tuning techniques such as LoRA, Adapter, and full fine-tuning in the sense that it preserves the integrity of the pre-trained knowledge. 

    \item We propose PT-PEFT, a method that sequentially performs the prefix-tuning followed by conventional fine-tuning technique, to maximize the utilization of pre-trained knowledge in LMMs. Our experimental results demonstrate that PT-PEFT outperforms the conventional fine-tuning methods in image captioning and VQA tasks.
    
\end{itemize}


\begin{figure*}[t!]
    \centering
    \begin{subfigure}[b]{0.32\textwidth}
        \centering
        \caption{\normalsize{Full Fine-tuning}}
        \includegraphics[width=\textwidth]{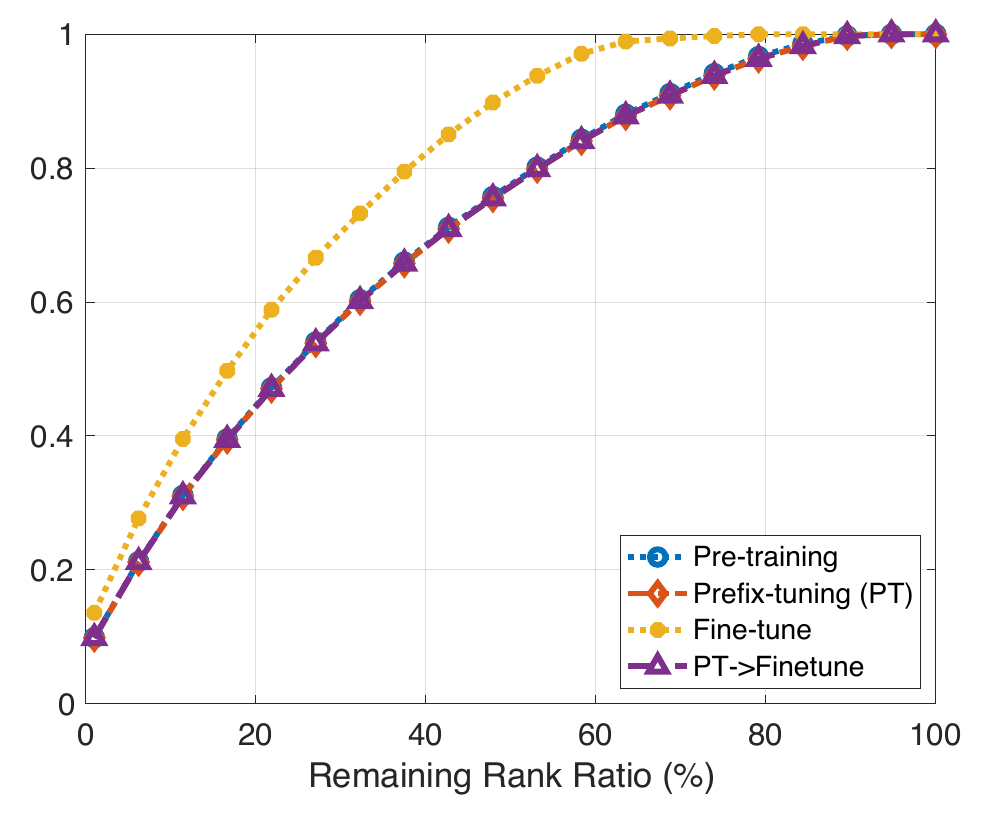}
        \label{fig:svd_ranks_vinvl}
    \end{subfigure}
    \hfill
    \begin{subfigure}[b]{0.32\textwidth}
        \centering
        \caption{\normalsize{LoRA}}
        \includegraphics[width=\textwidth]{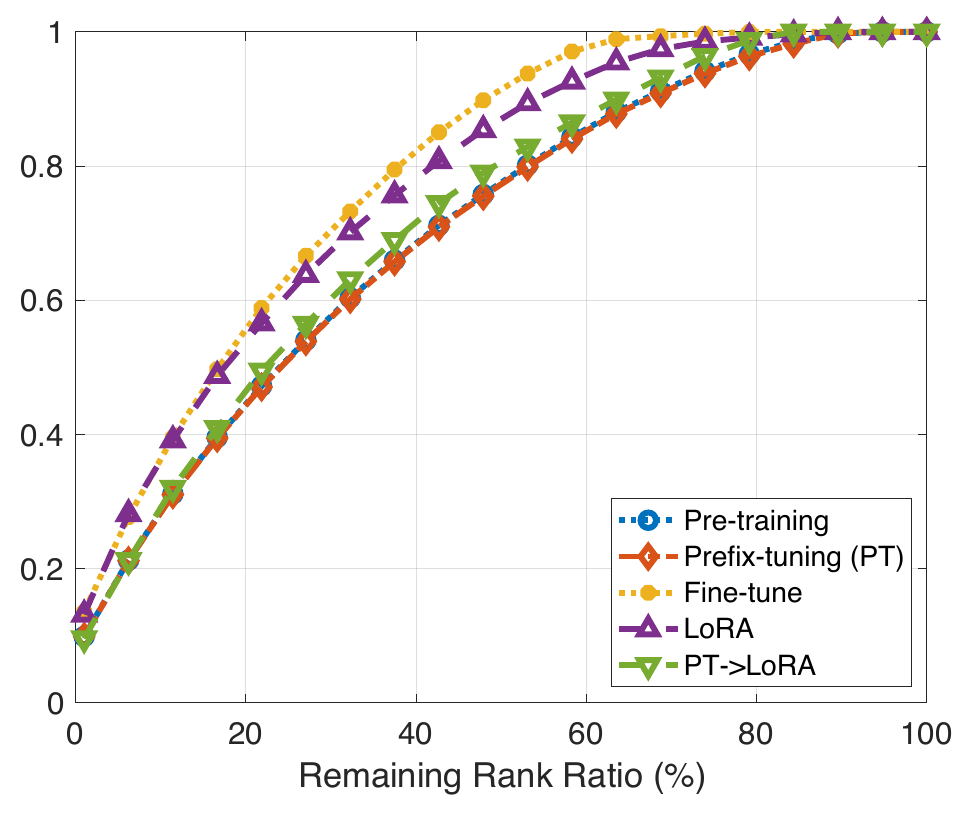}
        \label{fig:svd_ranks_blip}
    \end{subfigure}
    \hfill
    \begin{subfigure}[b]{0.32\textwidth}
        \centering
        \caption{\normalsize{S-Adapter}}
        \includegraphics[width=\textwidth]{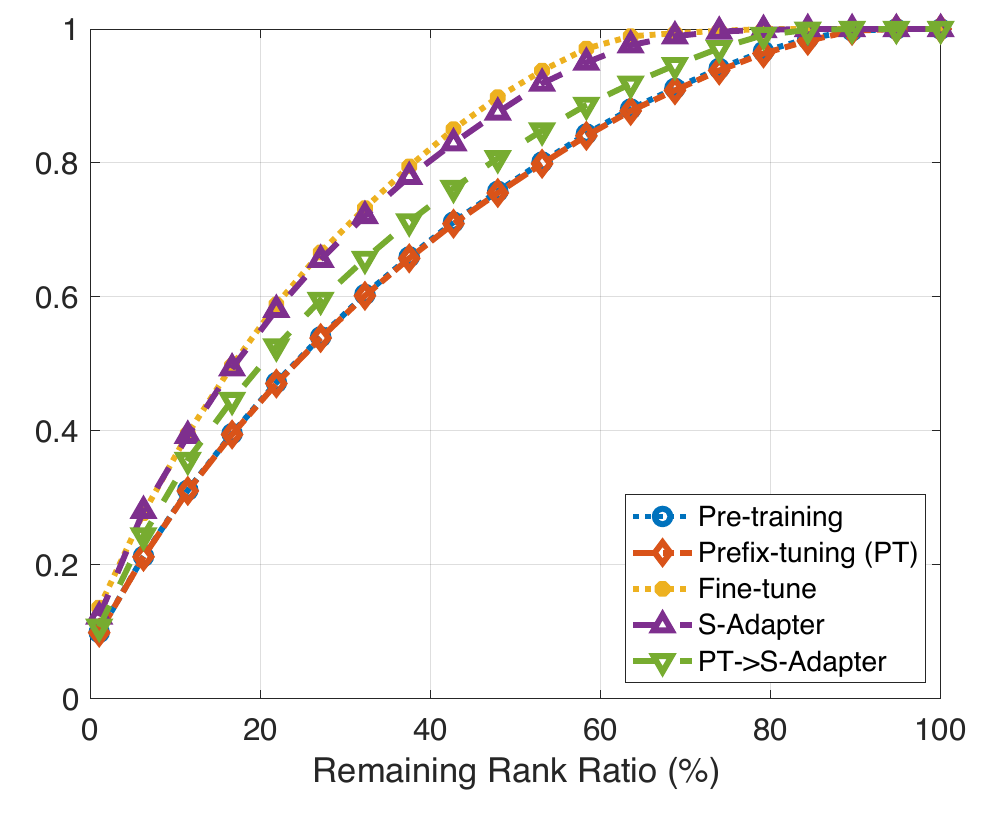}
        \label{fig:svd_ranks_blip}
    \end{subfigure}
    \vspace{-0.6cm}
    \caption{Accumulated and normalized singular values of features extracted from the last layer of BLIP-2. A more concave graph indicates that the singular values are more concentrated, implying the narrower representation space.}
    \label{fig:svd_ranks}
    \vspace{0.2cm}
\end{figure*}

\begin{table*}[t]

\vspace{-0.1cm}
\centering
\setlength\heavyrulewidth{1.5pt}
\renewcommand{\arraystretch}{1.2}
\resizebox{\textwidth}{!}{
\begin{tabular}{ccccccccccc}
\toprule
       & Pre-training & Fine-tuning & Prefix-tuning & S-Adapter & P-Adapter & LoRA   & PT$\xrightarrow{}$S-Adapter & PT$\xrightarrow{}$P-Adapter & PT$\xrightarrow{}$LoRA & PT $\xrightarrow{}$ Fine-tuning\\
\midrule
\midrule
VINVL  &  50.2 \% & 30.0 \% &   50.2  \% &    -  &    -  &  -  & - & - & -  & \textbf{50.2  \%}     \\
\midrule
BLIP-2 &     68.2 \% &  47.0  \% &    68.2  \% & 53.0 \% & 53.7 \% & 52.0 \%  & 63.5 \% & 58.4 \% & 63.5 \% & \textbf{68.2 \%} \\
\bottomrule
\end{tabular}}
\caption{Effective rank of representation space of various fine-tuning techniques. Note that the effective rank is defined as the remaining rank ratio at which the accumulated singular values equal to 0.9 in Figure~\ref{fig:svd_ranks}.}
\label{tab:effective_rank}
\end{table*}
\section{Representation Space Collapse Causes the Loss of Generalization Capabilities}
\label{sec:background}

\subsection{Zero-shot Sometimes Performs Better than Fine-tune}
In general, model parameter tuning performs better than the prefix-tuning. 
However, the full fine-tuned model generates even worse answers than the zero-shot generation for some cases.
Figure~\ref{fig:zero} presents a qualitative comparison between zero-shot inference, full fine-tuning, and prefix-tuning on IC and VQA tasks.
In IC tasks, we find that prefix-tuning is better than full fine-tuning in capturing detailed descriptions of objects.
Although the IC output from the fine-tuning is technically sound, captions generated through the prefix-tuning are rich in context and more natural.
Similarly, for VQA tasks, we observe that Top-5 answers from the prefix-tuning are more relevant to the given questions, whereas the answers from the fine-tuning are often irrelevant or less likely to be correct.

These results stem from the problem that the downstream dataset often lacks the object and attribute diversity compared to the dataset used for the pre-training.
Consequently, models may lose the learned word and image representations for objects and attributes during the fine-tuning.
This issue, known as catastrophic forgetting, undermines the model's ability to retain valuable pre-trained knowledge~\citep{rebuffi2017icarl, kalajdzievski2024scaling}.

\subsection{Relationship Between Semantic Richness and Representation Space}
In the vector space, catastrophic forgetting appears as the rank reduction of the representation matrix, so-called the \textit{representation collapse}.
The information contained within the representation matrix is closely associated with its rank~\citep{zhang2021orthogonality, bansal2018can, swaminathan2020sparse}. 
For instance, low-rank compression methods intentionally pursue a reduction in the rank of the feature matrix to extract essential information exclusively~\citep{sainath2013low, swaminathan2020sparse}.
Just as other information is expunged by low-rank compression, the representation collapse by catastrophic forgetting makes the representation matrix lose semantically rich details in objects and their attributes, potentially degrading the generalization ability for downstream tasks.

\subsection{Empirical Analysis on Representation Space Collapse}
\label{ssec:rank_svd}

\paragraph{Representation Space Analysis via SVD}
To quantitatively measure the representation collapse in different model adaptation methods, we apply SVD on the representation matrices.
SVD allows us to quantitatively analyze the average number of basis singular vectors used to represent a single text or image.
In our SVD analysis, we utilize the activation matrix of the model's last layer.
Specifically, LMM processes the text input $x_{txt}=\left[ w_{sos}, w_{1}, ... ,  w_{N},  w_{eos} \right]$, yielding a sequence of output embedding vectors $\text{F}_{txt}=\text{LMM}(x_{txt})$:
\begin{equation}
\label{eqn:wordembedding}
\text{F}_{txt}:= \left[ \textbf{f}_{txt}^{sos},  \textbf{f}_{txt}^{w_1},  ...,  \textbf{f}_{txt}^{w_N}, \textbf{f}_{txt}^{eos} \right].
\end{equation}
Using $\text{F}_{txt}$, we perform SVD and obtain the singular values (i.e., the diagonal elements of $\Sigma$):
\begin{equation}
\label{eqn:SVD}
\text{F}_{txt}=\text{U}\Sigma \text{V}^\text{T}.
\end{equation}
We sort the singular values $\textbf{s}=\left[ \sigma_1, ..., \sigma_M \right]$ in descending order and normalize such that sum of all singular values equals one:
\begin{equation}
\label{eqn:normalize}
\hat{\textbf{s}}=\frac{1}{\sum_{i=1}^{M}{\sigma_i}} \left[\sigma_1, ..., \sigma_M \right].
\end{equation}
After computing singular values on a per-image or per-sentence basis, we average them across the $K$ samples in the dataset:
\begin{equation}
\label{eqn:forallsamples}
\hat{\textbf{s}}_{avg}=\frac{1}{K}\sum_{i=1}^{K}{\hat{\textbf{s}}_k}.
\end{equation}
Finally, we compute the cumulative sum of the elements in $\hat{\textbf{s}}_{avg}$:
\begin{equation}
\label{eqn:accumulatedsum}
\textbf{y}=\left[\hat{\sigma}_{avg, 1}, ..., \sum_{j=1}^{i}{\hat{\sigma}_{avg, j}}, ..., \sum_{j=1}^{M}{\hat{\sigma}_{avg, j}} \right].
\end{equation}
The sum $\textbf{y}$ is plotted in Figure~\ref{fig:svd_ranks} for each model and training method.

\paragraph{Comparison Between Various Fine-tuning Methods}
Figure~\ref{fig:svd_ranks} presents the cumulative sum of singular values in feature matrices extracted from different models.
Specifically, we compare the rank of image and text features extracted from three distinct models (pre-trained, fine-tuned, and prefix-tuned).
The naive fine-tuned model shows the fastest saturation towards the top (see the red line in Figure~\ref{fig:svd_ranks}), meaning that most singular values are close to zero (i.e., $\sum_{i=1}^{k}\sigma_{i} \approx 1$ for small $k$). 
This in turn means that the effective rank of the feature matrix extracted from the fine-tuned model is much lower than that of the pre-trained model.

As shown in Table~\ref{tab:effective_rank}, LoRA-tuned and fine-tuned models utilize only 60\% of the basis vectors from the pre-trained representation space, while the prefix-tuning exploits almost all the basis vectors.
In addition, as shown in Figure~\ref{fig:svd_ranks}, the curvature of the singular value plot is highly correlated with final performance metrics (e.g., CIDEr, Accuracy)~\citep{daneshmand2020BN,dong2021loserank}.

\section{Prefix-Tuned Parameter-Efficient Fine Tuning (PT-PEFT)}
\label{subsec:method_imt}

\begin{figure}[t!]
    \centering
    \includegraphics[width=1.0\linewidth]{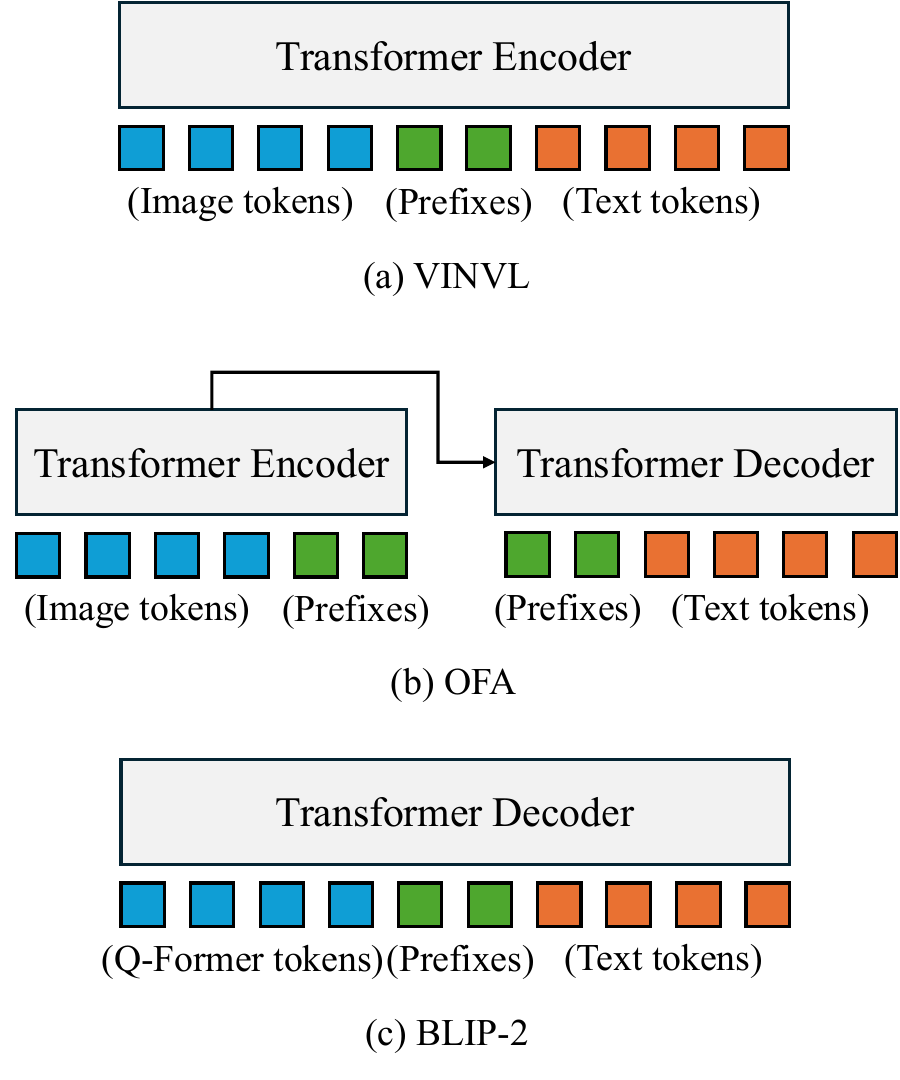}
    \caption{
        Visualization of where the prefixes are inserted for different LMMs. The proposed method can be applied for general Transformer-based architectures.}
    \label{fig:prefix_models}
\end{figure}

\paragraph{Prefix Implementation}
Prefix embedding vectors are first processed through the prefix encoder, following standard practices in prefix-tuning~\citep{li2021prefix} (see Appendix for details).
The processed prefixes are then concatenated with text and/or image tokens to form the input to the LMMs.
Figure~\ref{fig:prefix_models} illustrates various LMM architectures that can take prefixes as inputs.
The green boxes in the Figure represent learnable prefix embeddings (tokens) used during the prefix-tuning stage.

\paragraph{Two-stage Optimization}\label{subsubsec:two_stage}
We employ a two-stage approach: prefix-tuning followed by fine-tuning.
In the prefix-tuning stage, we only train the prefix embeddings and prefix encoder, keeping the other parameters of LMMs frozen.
In the fine-tuning stage, we adjust the parameters, including prefixes, to further adapt the model and enhance downstream performance.
Here, the parameters to be adjusted depend on whether it is PEFT or full fine-tuning.
\section{Experiments}\label{sec:result}
\setlength{\tabcolsep}{12pt}
\begin{table*}[t]

\vspace{-0.15cm}
\centering
\setlength\heavyrulewidth{1.5pt}
\renewcommand{\arraystretch}{1.0}
\resizebox{\linewidth}{!}{
\begin{tabular}{@{}lrcccccccc@{}}
\toprule
\multirow{2}{*}{} & \multirow{2}{*}{\#Trainable Params} & \multicolumn{3}{c}{COCO IC} & \multicolumn{3}{c}{Flickr30k IC} & \multicolumn{2}{c}{VQAv2} \\
\cmidrule(l){3-5}
\cmidrule(l){6-8}
\cmidrule(l){9-10}
& & B4 & C & S & B4 & C & S & test-dev & test-std \\
\midrule
\midrule
\multicolumn{10}{c}{OFA\textsubscript{BASE}~\cite{wang2022ofa}} \\
Prefix-tuning & 0.15 \% & 35.2 & 115.6 & 19.3 & 27.0 & 61.4& 16.5 & 72.9 & 73.2\\
\midrule
S-Adapter & 3.10\% &35.6  & 119.7 & 20.9 & 27.4 & 62.1 & 16.8 & 73.1 & 73.4\\
S-Adapter $\rightarrow$ \text{Prefix} &3.15\% & 38.2 & 128.2 & 21.6 & 27.6 & 64.8& 17.3 & 73.9 & 74.1\\
\textbf{Prefix} $\rightarrow$ \textbf{S-Adapter} & 3.15\% & \textbf{39.0} & \textbf{130.7} & \textbf{22.5} & \textbf{29.2} & \textbf{68.3} & \textbf{17.3} & \textbf{74.3} & \textbf{74.4} \\
\midrule
P-Adapter & 3.08\% & 36.8 & 123.7 & 21.3 & 28.5& 64.4 & 17.0&  73.4& 73.8\\
P-Adapter $\rightarrow$ \text{Prefix} & 3.12 \% & 38.4 & 129.7 & 21.7 & 28.8 & 67.2& 17.9 & 74.0 & 74.2\\
\textbf{Prefix} $\rightarrow$ \textbf{P-Adapter} &3.12 \% & \textbf{39.7} & \textbf{132.8} & \textbf{23.4} & \textbf{31.1} & \textbf{73.6} & \textbf{18.7} & \textbf{75.6} & \textbf{75.7} \\
\midrule
LoRA & 0.26 \% & 35.3 & 117.4 & 19.5 & 24.7& 52.4 & 15.2& 50.1 & 50.3\\
LoRA $\rightarrow$ \text{Prefix} & 0.45 \% & 36.6 & 122.0 & 21.2 & 28.5 & 66.2& 17.5 & 70.9 & 71.1\\
\textbf{Prefix} $\rightarrow$ \textbf{LoRA} &0.45 \% & \textbf{39.2} & \textbf{131.6} & \textbf{23.1} & \textbf{30.5} & \textbf{71.6} & \textbf{18.0} & \textbf{74.6} & \textbf{74.9} \\
\midrule
{Full fine-tuning} & 100 \% & 38.6 & 127.5 & {22.8}  & 32.2 & 74.1 & 18.5 & 75.7 & 75.8\\
\midrule \midrule
\multicolumn{10}{c}{BLIP-2\textsubscript{ViT-g + OPT 2.7B}~\cite{li2023blip2}} \\
Prefix-tuning & 0.20 \% & 41.0 & 138.0 & 24.9 & 34.6 & 92.3 & 20.6 & 30.1 & 29.8 \\
\midrule
S-Adapter & 4.32 \% & 40.4 & 140.0 & 25.0 & 34.4 & 93.8 & 22.6 & 51.8 & 52.4 \\
S-Adapter $\rightarrow$ \text{Prefix} & 4.52 \% & 40.7 & 139.8 & 24.8 & 34.9 & 93.8 & 22.7 & {53.2} & 54.3 \\
\textbf{Prefix} $\rightarrow$ \textbf{S-Adapter} & 4.52 \% & \textbf{41.0} & \textbf{140.6} & \textbf{25.0} & \textbf{35.6} & \textbf{95.4} & \textbf{23.4} & \textbf{54.3} &\textbf{54.4} \\
\midrule
P-Adapter & 3.23 \% & 40.1 & 139.0 & 24.9 & 33.6 & 90.4 & 22.3 & 53.1 & 50.4 \\
P-Adapter $\rightarrow$ \text{Prefix} & 3.43 \% & 40.6 & 140.6 & 24.9 & 35.0 & 94.1 & 23.0 & {53.2} &\ 53.7 \\
\textbf{Prefix} $\rightarrow$ \textbf{P-Adapter} & 3.43 \% & \textbf{41.0} & \textbf{140.6} & \textbf{25.2} & \textbf{35.1} & \textbf{95.1} & \textbf{23.4} & \textbf{53.2} &\textbf{54.3} \\
\midrule
LoRA & 0.34 \% & 40.3 & 139.0 & 25.1 & 35.2 & 94.4 & 22.5 & 43.8 & 44.4 \\
LoRA $\rightarrow$ \text{Prefix} & 0.54 \% & {40.6} & {139.3} & {25.0} & {35.7} & {95.9} & {23.0} & {53.2} &{54.3} \\
\textbf{Prefix} $\rightarrow$ \textbf{LoRA} & 0.54 \% & \textbf{41.2} & \textbf{140.6} & \textbf{25.2} & \textbf{36.1} & \textbf{97.0} & \textbf{23.3} & \textbf{52.2} &\textbf{52.3} \\
\midrule
Full fine-tuning & 100 \% & 41.1 & 141.7 & 25.0  & 35.9 & 97.5 & 27.6 & 74.9 & 74.7 \\
\bottomrule
\end{tabular}
}
\caption{Performance comparison between PEFT and our PT-PEFT, applying prefix-tuning followed by other PEFT. B4, C, and S indicate BLEU-4, CIDEr, and SPICE scores, respectively.}
\label{tab:comp_lora}
\vspace{-0.1cm}
\end{table*}
\setlength{\tabcolsep}{6pt}
\subsection{Setup}\label{subsec:setup}

\paragraph{Model}
To demonstrate the generalization capability of our method, we evaluate various pre-trained LMMs with different architectures and sizes.
Specifically, we conduct experiments on VINVL-BASE/LARGE~\citep{zhang2021vinvl}, OFA-BASE~\citep{wang2022ofa}, BLIP~\citep{li2022blip}, an BLIP-2~\citep{li2023blip2} models.

\paragraph{Dataset}
We evaluate image captioning (IC) task performance on MS-COCO~\citep{lin2014coco} and Flickr30k~\citep{plummer2015flickr30k} datasets.
For the visual question-answering (VQA) task, we use the VQAv2~\citep{antol2015vqa} dataset.

\paragraph{Fine-tuning Methods}
We take pre-trained LMMs and compare different fine-tuning methods.
These include Prefix-tuning (Prefix), LoRA, Parallel-Adapter (P-Adapter), and Sequential-Adapter (S-Adapter)~\cite{hu2023llm}, and also the full fine-tuning (Full-FT).
Adapters usually include multi-layer modules, therefore they generally equip more trainable parameters than LoRA.
Prefix-tuning uses the smallest number of trainable parameters among all.
For fair comparison across PEFT methods, we matched the number of trainable parameters.
Note that our PT-PEFT can be applied to all methods, with prefix-tuning used before other fine-tuning methods as our key innovation.

\paragraph{Additional Details}
We carefully designed settings for each model and method to achieve the best performance.
For more details about the models, datasets, and hyper-parameters, please refer to Appendix~\ref{supp:sec:detail}.

\subsection{Downstream Task Performance}\label{subsec:performance_captioning}

\paragraph{Prefix-tuned PEFT}
Table~\ref{tab:comp_lora} shows the performance of various task adaptation methods, applied to OFA-BASE and BLIP-2 models.
Our proposed PT-PEFT consistently outperforms standard PEFT methods across all 8 metrics.
PT-PEFT even surpasses full fine-tuning, with a 0.2p/0.1p in BLEU-4 metric for Flickr30k/COCO, along with a 0.2p improvement in SPICE score.
Additionally, the results show that applying PEFT before prefix-tuning (i.e., reversing the order) is considerably less effective than PT-PEFT, though it still performs better than not using prefix-tuning at all.

\setlength{\tabcolsep}{14pt}
\begin{table*}[t]

\vspace{-0.15cm}
\centering
\setlength\heavyrulewidth{1.5pt}
\renewcommand{\arraystretch}{1.0}
\resizebox{1.0\textwidth}{!}{
\begin{tabular}{@{}lccccccc@{}}
\toprule
\multirow{2}{*}{}       & 
\multicolumn{3}{c}{COCO Image Captioning} & 
\multicolumn{3}{c}{Flickr-30k Image Captioning} & 
\\ 
\cmidrule(l){2-4} 
\cmidrule(l){5-7}
& BLEU-4 & CIDEr & SPICE & BLEU-4 & CIDEr & SPICE \\ \midrule \midrule
& \multicolumn{6}{c}{VINVL$_{\text{BASE}}$}
\\
{Prefix-tuning} & 37.3  & 122.5 & 22.2 & 28.7 & 65.5 & 16.9
\\ 
{Full fine-tuning} & 40.4  & 137.2 & 24.5 & 33.8 & 85.5 & 21.1
\\ 
{\textbf{Prefix $\rightarrow$ Full-FT}} & \textbf{41.2 $\pm$ 0.08} & \textbf{141.1 $\pm$ 0.10} & \textbf{25.0 $\pm$ 0.04} & \textbf{35.6 $\pm$ 0.13} & \textbf{89.7 $\pm$ 0.36} & \textbf{21.5 $\pm$ 0.10}  
\\ 

\midrule
& \multicolumn{6}{c}{VINVL$_{\text{LARGE}}$}
\\
{Prefix-tuning} & 38.5 & 128.2 & {23.2}  & 31.9 & 72.0 & 18.3
\\ 
{Full fine-tuning} & 41.0 & 139.6 & {24.8}  & 34.3 & 85.2 & 21.1
\\ 
{\textbf{Prefix $\rightarrow$ Full-FT}} & \textbf{41.4 $\pm$ 0.06} & \textbf{141.1 $\pm$ 0.12} & \textbf{24.9 $\pm$ 0.07} & \textbf{35.8 $\pm$ 0.59} & \textbf{89.8 $\pm$ 0.14} & \textbf{21.9 $\pm$ 0.04}
\\

\midrule
& \multicolumn{6}{c}{OFA$_{\text{BASE}}$}
\\

{Zero-shot} & 18.2 & 62.3 & {14.8}  & 15.3 & 23.2 & 12.1
\\ 

{Prefix-tuning} & 35.2 & 115.6 & {19.3}  & 27.0 & 61.4 & 16.5
\\ 

{Full fine-tuning} & 38.6 & 127.5 & {22.8}  & 32.2 & 74.1 & 18.9
\\ 

{\textbf{Prefix $\rightarrow$ Full-FT}} & \textbf{41.4 $\pm$ 0.02} & \textbf{136.4 $\pm$ 0.16} & \textbf{24.3 $\pm$ 0.11} & \textbf{35.8 $\pm$ 0.24} & \textbf{89.8 $\pm$ 0.21} & \textbf{21.9 $\pm$ 0.07}
\\
\midrule
& \multicolumn{6}{c}{BLIP-2$_{\text{ViT-g + OPT 2.7B}}$}

\\
{Zero-shot} & 39.7 & 129.0 & 22.6  & 29.5 & 74.5 & 16.8
\\ 
{Prefix-tuning} & 40.0 & 138.0 & 24.9  & 34.6 & 92.3 & 20.6
\\ 
{Full fine-tuning} & 41.1 & 141.7 & 25.0  & 35.9 & 97.5 & \textbf{27.6}
\\ 
{\textbf{Prefix $\rightarrow$ Full-FT}} & \textbf{41.8 $\pm$ 0.11} & \textbf{142.8 $\pm$ 0.07} & \textbf{25.2 $\pm$ 0.04} & \textbf{36.5 $\pm$ 0.09} & \textbf{98.3 $\pm$ 0.19} & 23.6 $\pm$ 0.30
\\

\bottomrule
\end{tabular}
}
\caption{
Image captioning performance comparison between prefix-tuning, full fine-tuning and ours.
}
\label{tab:coco_metric}
\end{table*}
\setlength{\tabcolsep}{6pt}

\setlength{\tabcolsep}{23pt}
\begin{table}[t]
\centering
\setlength\heavyrulewidth{1.5pt}
\renewcommand{\arraystretch}{1.0}
\resizebox{1.0\linewidth}{!}{
\begin{tabular}{@{}lcc@{}}
\toprule
\multirow{2}{*}{} & \multicolumn{2}{c}{VQAv2} \\
\cmidrule(l){2-3}
& test-std & test-dev \\
\midrule
\midrule
& \multicolumn{2}{c}{VINVL\textsubscript{BASE}} \\
Linear-probing & 72.7 & 72.6 \\
Prefix-tuning & 73.8 & 73.4 \\
Full fine-tuning & 74.1 & 74.4 \\
\textbf{Prefix $\rightarrow$ Full-FT} & \textbf{76.2 $\pm$ 0.04} & \textbf{76.2 $\pm$ 0.08} \\
\midrule
& \multicolumn{2}{c}{VINVL\textsubscript{LARGE}} \\
Linear-probing & 73.3 & 73.7 \\
Prefix-tuning & 75.0 & 74.9 \\
Full fine-tuning & 76.5 & 76.6 \\
\textbf{Prefix $\rightarrow$ Full-FT} & \textbf{77.0 $\pm$ 0.04} & \textbf{77.9 $\pm$ 0.02} \\
\midrule
& \multicolumn{2}{c}{OFA\textsubscript{BASE}} \\
Zero-shot & 25.9 & 25.8 \\
Prefix-tuning & 73.2 & 72.9 \\
Full fine-tuning & 75.8 & 75.7 \\
\textbf{Prefix $\rightarrow$ Full-FT} & \textbf{76.8 $\pm$  0.04} & \textbf{76.6 $\pm$ 0.04} \\
\midrule
& \multicolumn{2}{c}{BLIP\textsubscript{LARGE}} \\
Zero-shot & 5.0 & 5.2 \\
Prefix-tuning & 30.1 & 29.8 \\
Full fine-tuning & 74.9 & 74.7 \\
\textbf{Prefix $\rightarrow$ Full-FT} & \textbf{77.0 $\pm$ 0.07} & \textbf{77.9 $\pm$ 0.03} \\
\bottomrule
\end{tabular}
}
\caption{VQAv2 performance comparison.}
\label{tab:vqa_metric}
\vspace{-0.2cm}
\end{table}
\setlength{\tabcolsep}{6pt}

\paragraph{Prefix-tuned Full Fine-tuning}
Tables~\ref{tab:coco_metric} and~\ref{tab:vqa_metric} compare prefix-tuning, full fine-tuning, and the sequential combination of both (ours).
To ensure the reliability of our results, we conducted three separate runs with different random seeds and reported the mean and standard deviation obtained from these runs.
Notably, the standard deviation of the scores is significantly smaller than the improvements over the baseline models.
Compared to the full fine-tuning, our prefix-tuned full fine-tuning achieves approximately an 11\% increase in the BLEU-4, a 16\% increase in SPICE, and a noteworthy 21\% improvement in CIDEr. 
These results highlight the effectiveness of our method, demonstrating that prefix-tuning can help preserve pre-trained knowledge and improve performance in both PEFT and full fine-tuning scenarios.

\setlength{\tabcolsep}{10pt}
\begin{table}[ht]
    \setlength\heavyrulewidth{1.5pt}
    \renewcommand{\arraystretch}{1.0}
 
    \begin{subtable}{0.48\textwidth}
    \centering
    \resizebox{\linewidth}{!}{
    \begin{tabular}{lccccc}
        \toprule
        \multirow{2}{*}{} & \multicolumn{3}{c}{COCO IC valid} & \multicolumn{2}{c}{VQAv2 valid}           \\
        \cmidrule(l){2-4} \cmidrule(l){5-6}
        & B4 & C & S & Acc1 & \multicolumn{1}{c}{Acc5} \\
        \midrule
        \midrule
        w/Prefix & 41.3 & 139.3 & 24.6 & 75.2 & 93.3 \\
        \midrule
        -Prefix & 22.9 & 75.0 & 15.3 & 36.5 & 72.6 \\
        \midrule
        -Prefix +Noise & 25.1 & 82.9 & 16.2 & 31.2 & 61.4 \\
        \bottomrule
    \end{tabular}
    }
    \caption{Performance of the sequential-tuned model.}
    \vspace{0.2cm}
    \end{subtable}
    \vfill
    \begin{subtable}{0.48\textwidth}
    \centering
    \resizebox{\linewidth}{!}{
    \begin{tabular}{lccccc}
        \toprule
        \multirow{2}{*}{} & \multicolumn{3}{c}{COCO IC valid} & \multicolumn{2}{c}{VQAv2 valid}           \\
        \cmidrule(l){2-4} \cmidrule(l){5-6}
        & B4 & C & S & Acc1 & \multicolumn{1}{c}{Acc5} \\
        \midrule
        \midrule
        w/Prefix & 41.0 & 138.0 & 24.3 & 71.6 & 91.9 \\
        \midrule
        -Prefix & 23.1 & 74.3 & 15.1 & 72.2 & 91.7 \\
        \midrule
        -Prefix +Noise & 23.5 & 76.8 & 15.5 & 62.2 & 86.6 \\
        \bottomrule
    \end{tabular}
    }
    \caption{Performance of the parallel-tuned model.}
    \vspace{0.1cm}
    \end{subtable}
    \caption{Comparison of (a) sequential and (b) parallel tuning. Unlike PT-PEFT, parallel tuning applies prefix-tuning and fine-tuning together. For noise addition experiments (third rows), we replace learned prefixes with random noise during inference.}
    \label{tab:ablation_prefix}
     \vspace{-0.2cm}
\end{table}
\setlength{\tabcolsep}{6pt}
\setlength{\tabcolsep}{11pt}
\begin{table}[t]

\vspace{-0.15cm}
\setlength\heavyrulewidth{1.5pt}
\renewcommand{\arraystretch}{1.0}
\centering
\resizebox{1.0\linewidth}{!}{
\begin{tabular}{cccccc}
\toprule
\multirow{2}{*}{{Model}} & \multicolumn{2}{c}{\#Epochs} & \multicolumn{3}{c}{{COCO Image Captioning}} \\
                                &              PT                                   &     FT                                    & BLEU-4           & CIDEr          & SPICE           \\
\midrule
\midrule
M1                              & 3                                             & 7                                           & 35.3             & 114.2          & 18.8           \\
\midrule
M2                              & 5                                             & 5                                           & 40.2             & 129.6          & 23.5           \\
\midrule
M3                              & 7                                             & 3                                           & 41.4             & 136.4          & 24.3          \\ \bottomrule
\end{tabular}
}
\caption{Ablation study on the number of epochs for prefix-tuning (PT) and fine-tuning (FT) stages.}
\label{tab:training_steps}
\end{table}
\setlength{\tabcolsep}{6pt}

\section{Analysis \& Discussion}
\label{sec:analysis}

\subsection{Preserving Representation Space}
Figure~\ref{fig:svd_ranks} visualizes the accumulated singular values, as described in Section~\ref{ssec:rank_svd}.
The saturation curves for the pre-trained, prefix-tuned, and PT-PEFT models are almost identical, implying that the effective rank is preserved after training.
In contrast, LoRA, Adapter, and full fine-tuning methods show more concave curves, indicating a narrower representation space.

\subsection{Ablation Study}

\paragraph{Sequential vs. Parallel}
Instead of sequentially applying prefix-tuning and then fine-tuning, one may consider using both methods together in parallel.
We call this variant \textit{parallel-tuning} and compare its performance to our sequential training.
Table~\ref{tab:ablation_prefix}~(a) and~(b) present the downstream task performance of parallel tuning and ours, respectively.
The result shows that parallel-tuning performs worse than PT-PEFT in all cases.

To further investigate how parallel-tuning affects the effectiveness of the prefix, we distort the trained prefixes and observe the performance change.
Table~\ref{tab:ablation_prefix}(b) shows that for the parallel-tuned model, even without prefixes, VQA accuracy is almost preserved, meaning that the prefix does not contribute to performance.
This finding is further emphasized when replacing the trained prefix with random noise; accuracy only slightly decreases, implying that the prefixes are not very powerful.
In contrast, when using prefix tuning first (Table~\ref{tab:ablation_prefix}(a)), removing prefixes severely hurts the accuracy, showing that they actively contribute to the performance.

\paragraph{Ratio of Each Stage}
We conduct experiments to find the best number of training steps for the prefix-tuning and fine-tuning stages.
As shown in Table~\ref{tab:training_steps}, we found that prefix-tuning requires a sufficiently long iteration for optimal performance.
Within the same training budget, the model achieves better performance with fewer fine-tuning epochs if sufficient prefix-tuning precedes.

\subsection{Intuitive Explanation of PT-PEFT}
Based on the analysis, we conclude that prefix-tuning and other fine-tuning methods contribute to the adaptation in different ways.
By sequentially performing prefix-tuning and parameter fine-tuning, the model first encodes the representation space as prefix tokens that align with the pre-trained space.
This is because the original model parameters remain unchanged during prefix-tuning, so the learned knowledge is not damaged.
Once such context is established, the subsequent fine-tuning process can effectively avoid the representation collapse, as the prefixes provide a foundation for a rich representation space.

\subsection{Prior Works in Language Domain}
In this subsection, we highlight how our work differs from recent studies that combine two fine-tuning techniques in the language domain.
The original LoRA paper reported that combining LoRA with Prefix-tuning could improve performance (Appendix E of the paper~\citep{hu2021lora}).
However, their combination used a "parallel-tuning" approach, in contrast to our "sequential-tuning" approach.
In addition, they utilized a much larger number of trainable parameters, making it an unfair comparison between LoRA alone and LoRA with Prefix-tuning.

Around the same time as our work, ProMot~\citep{wang2024twostage} also suggested using prefix-tuning before model parameter tuning in a sequential manner.
They also reported significant performance improvements, which is consistent with our findings.
However, our work is very distinct in two key perspectives.

First, our experiments focus on LMMs, demonstrating the effectiveness of PT-PEFT across various vision-language tasks and Transformer-based model architectures.
Second, our analyses show that the primary reason for performance gain comes from the preservation of learned knowledge during pre-training, as revealed by our systematic investigation of the effective rank of embeddings.
This sets our work apart and highlights the uniqueness of our PT-PEFT.
\section{Conclusion}
\label{sec:conclusion}
In this paper, we discovered that fine-tuning methods including LoRA, Adapter, and full fine-tuning could cause the loss of learned knowledge from the pre-training stage.
We quantified this loss in representation space using a novel rank-based analysis and identified that prefix-tuning does not cause this critical loss.
Based on these findings, we proposed a two-step strategy, PT-PEFT, which first performs prefix-tuning and then applies other fine-tuning methods.
Our experiments showed that PT-PEFT not only preserves the representation space preservation but also improves downstream task performance.
\pagebreak
\section{Limitations}
\label{sec:limitation}
The proposed PT-PEFT can take advantage of both prefix-tuning and fine-tuning.
However, there are two practical limitations.
Firstly, it leads to an increased computational cost during inference due to the longer input sequence.
Managing this increased computational cost in prefix-tuning may become challenging, especially when the portion of prefixes in the total number of input tokens is large.
It's worth noting that the performance gains tend to plateau at around 16 prefixes, which doesn't significantly exacerbate the computational cost (see Appendix~\ref{sec:additional_analysis}, prefix length ablation study).
Secondly, we manually determine the best-performing hyper-parameters, such as prefix length, learning rates, and training iterations.
We did our best to find the best set for a fair comparison; however, we are aware that such a manual hyperparameter tuning process can be cumbersome, especially when applying our technique to new tasks, datasets, or models.

\section{Ethical Statement}
\label{sec:Ethics}
In our paper, we analyze various fine-tuning strategies to identify methods for preserving pre-trained knowledge during the fine-tuning process.
Rather than having potential risks, we believe that our research can serve as a solution to address ethical issues related to data corruption and safety control in current AI systems.
For instance, even if the model is fine-tuned with data corrupted by hacking, our technique can offer robustness to such data corruption by preserving the model’s representation space.
Our work can be also beneficial for not forgetting the safety guardrails learned during pre-training or instruction tuning. 
We'd like to note that this representation-preserving have not been studied much in VL models, regardless of the increasing interest on VL applications.

\section*{Acknowledgments}
This work was supported by the National Research Foundation of Korea(NRF) grant funded by the Korea government(MSIT)(RS-2023-00208985) and the Institute of Information \& communications Technology Planning \& Evaluation (IITP) under the Artificial Intelligence Semiconductor Support program to nurture the best talents (IITP-2023-RS-2023-00256081) grant funded by the Korea government(MSIT).
The authors would like to thank Jinwoo Son for his valuable assistance in proofreading this manuscript.


\bibliography{egbib}

\clearpage
\appendix
\begin{table*}[t]

\setlength\heavyrulewidth{1.5pt}
\renewcommand{\arraystretch}{1.0}
\resizebox{1.0\linewidth}{!}{
\begin{tabular}{cccccc}
\toprule
Model                              & \# of Param           & Module                                   & Hidden Dim & Number of Layer & Number of Attention Head \\ \midrule \midrule
VINVL Base                         & 110M                  & VL Fusion Encoder (BERT-Base)            & 768        & 12              & 12                       \\ \midrule
VINVL Large                        & 340M                  & VL Fusion Encoder (BERT-Large)           & 1024       & 24              & 16                       \\ \midrule
\multirow{3}{*}{OFA Base}          & \multirow{3}{*}{180M} & Vision Encoder (ResNet-101)              & 2048       & 101             & -                        \\ \cmidrule(){3-6}
                                   &                       & VL Fusion Encoder (Transformer Enc Base) & 768        & 6               & 12                       \\ \cmidrule(){3-6}
                                   &                       & VL Fusion Decoder (Transformer Dec Base) & 768        & 6               & 12                       \\ \midrule
\multirow{3}{*}{BLIP-2 (OPT 2.7B)} & \multirow{3}{*}{3.6B} & Vision Encoder (ViT-g)                   & 1408       & 40              & 16                       \\ \cmidrule(){3-6}
                                   &                       & Q-Former (BERT-Base)                     & 768        & 12              & 12                       \\ \cmidrule(){3-6}
                                   &                       & VL Fusion Decoder (OPT 2.7B)             & 2560       & 32              & 32                     \\ \bottomrule
\end{tabular}
}
\caption{Baseline VL pre-trained models specifications.}
\label{tab:baselines}
\end{table*}
\begin{table*}[ht]

\centering
\setlength\heavyrulewidth{1.5pt}
\renewcommand{\arraystretch}{1.0}
\resizebox{0.8\linewidth}{!}{
\begin{tabular}{cccc}
\toprule
Model                              & Module                                   & \begin{tabular}[c]{@{}c@{}}Prefix-tuning \\ Prefix Length\end{tabular} & \begin{tabular}[c]{@{}c@{}}LoRA \\ Weights\end{tabular} \\ \midrule \midrule
VINVL Base                         & VL Fusion Encoder (BERT-Base)            & 16                                                                     & -                                                       \\ \midrule
VINVL Large                        & VL Fusion Encoder (BERT-Large)           & 16                                                                     & -                                                       \\ \midrule
\multirow{3}{*}{OFA Base}          & Vision Encoder (ResNet-101)              & -                                                                      & -                                    \\ \cmidrule{2-4}
                                   & VL Fusion Encoder (Transformer Enc Base) & 64 (IC), 16 (VQA)                                                      & Q, K, V (r=16, a=32)                                    \\ \cmidrule{2-4}
                                   & VL Fusion Decoder (Transformer Dec Base) & 64 (IC), 16 (VQA)                                                      & Q, K, V (r=16, a=32)                                    \\ \midrule
\multirow{3}{*}{BLIP-2 (OPT 2.7B)} & Vision Encoder (ViT-g)                   & -                                                                      & Q, K, V (r=16, a=32)                                    \\ \cmidrule{2-4}
                                   & Q-Former (BERT-Base)                     & 8 (IC), 16 (VQA)                                                       & Q, K, V (r=16, a=32)                                    \\ \cmidrule{2-4}
                                   & VL Fusion Decoder (OPT 2.7B)             & 8 (IC), 16 (VQA)                                                       & -                   \\ \bottomrule
\end{tabular}
}
\caption{Parameter-efficient tuning (Prefix-tuning and LoRA) specifications.}
\label{tab:peft-params}
\vspace{-0.1cm}
\end{table*}
\section{Related Work}
\label{sec:related_work}


\paragraph{VL Model Architecture}
The Transformer and its variants (e.g., BERT, GPT) are widely adopted as VL model architectures due to their powerful attention mechanisms capturing correlations between image and text~\citep{vaswani2017attention}. Examples include VINVL using a Transformer encoder, OFA employing a Transformer encoder-decoder pair, and BLIP-2 utilizing a Transformer decoder. We evaluate PT-PEFT on these models to demonstrate its robustness and applicability.

\paragraph{VL Unsupervised Pre-training}
VL models often undergo unsupervised pre-training on large datasets, employing objectives like masked language modeling, image-text matching, and causal language modeling~\citep{li2023blip2, alayrac2022flamingo, wang2022ofa, yuan2021florence, zhang2021vinvl}. 
This pre-training helps the model understand the relationships between image and text. 
Tasks include predicting masked words, scoring image-text matching, and predicting the next words from given image-text pairs.

\paragraph{Semantic Richness and Rank}
Assessing the semantic richness of features is crucial for effective vision-language (VL) learning. 
This refers to how well a feature encapsulates fine-grained, dense information from the input. 
Evaluation includes linear probing in computer vision. Numerous studies indicate a strong correlation between rank and information content in representations~\citep{bansal2018can, zhang2021orthogonality}.
For instance, low-rank compression methods intentionally reduce rank to distill essential information, such as object class~\citep{sainath2013low, swaminathan2020sparse}.

\paragraph{Fine-tuning Strategies in VL Learning}
To enhance pre-trained model performance for downstream tasks, various transfer learning techniques address domain adaptation challenges.
A parameter-efficient fine-tuning approach often inserts additional modules into pre-trained model layers and optimizes only these modules~\citep{houlsby2019adapter, hu2021lora}.
Such PEFT methods are beneficial for greatly reducing the training cost by minimizing the number of trainable parameters.

\section{Experiments Setup}\label{supp:sec:detail}


\subsection{Model}
\paragraph{Baselines} To assess the effectiveness of PT-PEFT, we have employed a diverse set of pre-trained models featuring different architectures and sizes.
Specifically, we have tested models such as VINVL base, VINVL large~\citep{zhang2021vinvl}, OFA (base)~\citep{wang2022ofa}, BLIP~\citep{li2022blip}(only for VQA) and BLIP-2 (ViT-g and OPT-2.7B)~\citep{li2023blip2} as our baseline model due to its good performance on VL sequence generation and classification among many VL model variants~\citep{tan2019lxmert,lu2019vilbert,li2020visualbert,zhou2020unified,li2020oscar, alayrac2022flamingo}, as described in Table~\ref{tab:baselines}~\citep{zhang2021vinvl,wang2022ofa,li2023blip2}.

\begin{figure}[h]
\centering
\includegraphics[width=0.85\linewidth]{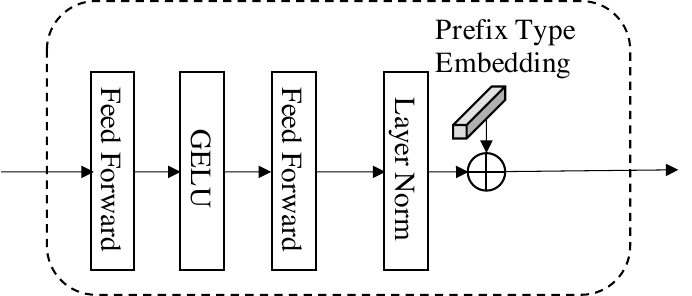}
\vspace{0.1cm}
\caption{Prefix encoder structure.}
\label{fig:imt_encoder}
\vspace{-0.5cm}
\end{figure}
\paragraph{Prefix Encoder} Figure~\ref{fig:imt_encoder} illustrates the prefix encoder (see Section~3).
In contrast to previous re-parameterizations~\citep{li2021prefix}, our approach incorporates prefix type embedding to establish a symmetrical setting with token type embedding, as used in previous VL models~\citep{zhang2021vinvl, li2020oscar}.
After training, the output of the prefix encoder can be saved as the new prefix, so there is no computational overhead in using this block.
In other words, the block is only realized during the training phase.

\subsection{Downstream task}

\paragraph{Visual Question Answering} Visual Question Answering task requires the model to select or generate the correct answer from the given question-image pair.
For VINVL~\citep{zhang2021vinvl,li2020oscar}, we train the model to classify the answer given question and image pair sequence from answer sets (i.e., 3129 for VQAv2, 1852 for GQA).
For OFA~\citep{wang2022ofa} and BLIP-2~\citep{li2023blip2}, we train the model to generate the answer given question and image pair.

\paragraph{Image Captioning} Image captioning task requires the model to generate a natural language description for the given input image.
Image captioning fine-tuning typically follows a 2-stage process, which consists of cross-entropy (CE) training and self-critical sequence training (SCST)~\citep{rennie2017scst}.

During CE training, the model uses CE loss to predict the correct words given image.
Then, the model is further trained by optimizing the CIDEr score with SCST which utilizes the score as the reward for REINFORCE algorithm~\citep{rennie2017scst}.
For inference, we utilize a beam size of 5 for beam search. 

\subsection{Dataset}
\paragraph{Image Captioning}
For IC experiments, we evaluate the performance of our proposed fine-tuning techniques on MS COCO~\citep{lin2014coco} and Flickr30k~\citep{plummer2015flickr30k} datasets.
We follow the Karpathy split~\citep{karpathy2015deep} for a fair comparison.
Karpathy split of COCO and Flickr30k datasets contain 83k/5k/5k and 29.8k/1k/1k images for train/val/test split.

\paragraph{Visual Question Answering}
For VQA experiments, the model is evaluated on the VQAv2 dataset~\citep{antol2015vqa}.
VQAv2 dataset contains 83k/41k/81k images and 444k/214k/448k question sets for train/val/test split, respectively.

\subsection{Experiment Details}

\paragraph{Hyper-parameters}
For training, we employ a set of hyper-parameters as detailed in Table~\ref{tab:hyper_param}. 
The table shows the best configurations for prefix-tuning and fine-tuning; these settings are also used for each stage of PT-PEFT. 
To update the network parameters, we utilize the AdamW optimizer \citep{loshchilov2017adamw} with betas set to (0.9, 0.99). 
For the learning rate scheduling, We combine linear warm-up followed by linear decay, gradually increasing the learning rate from 0 to the maximum LR during warm-up epochs and linearly decaying it to 0 for the remaining training epochs.


\paragraph{Evaluation Metrics}
In evaluating image captioning, we employ the CIDEr, SPICE, and BLEU-4 metrics \citep{vedantam2015cider, anderson2016spice, papineni2002bleu} to evaluate the quality of generated captions.
The evaluation is performed using the \texttt{pycocoevalcap} API available at \url{https://github.com/salaniz/pycocoevalcap}.
For visual question answering, we present accuracy as a performance metric.

\paragraph{Computational Resources}
We conducted experiments using four A100 (40GB) GPUs.

\subsection{Implementation Details}
\paragraph{Prefix-tuning} 
In prefix-tuning, the VL model is kept frozen, and only the prefix-encoder block (see Figure~\ref{fig:imt_encoder}) and prefix vectors are trained. 
Our implementation of the prefix-tuning closely follows the original prefix-tuning approach~\citep{li2021prefix}, where an MLP is employed as the prefix encoder for stable optimization.
The number of prefix vectors is empirically chosen for the best performance based on the experiment in Figure~\ref{fig:prefix_len_ablation} as described in Table~\ref{tab:peft-params}.

\paragraph{LoRA}
We implement the low-rank adapter following~\citep{hu2021lora}.
We update all query, key, and value projection matrices in the self-attention module by setting the rank $r=16$, scaling factor $\alpha=32$, and dropout probability of 0.05 throughout all experiments (see Table~\ref{tab:peft-params}).
 
\paragraph{PT-PEFT}
For image captioning, we freeze the word embedding layer and the head throughout the training process, including both the prefix-tuning stage and the subsequent fine-tuning stage. 
In the prefix-tuning stage, we only train the prefix encoder and prefix embedding using CE training. 
Subsequently, we fine-tune the model using a combination of CE training and SCST (for VINVL COCO-IC only).
For visual question answering, we follow a similar procedure.
We first train the prefix encoder and prefix embedding (and the CLS head for VINVL) and then proceed with fine-tuning the model.

\paragraph{PT-LoRA}
PT-LoRA is the parameter-efficient version of prefix-tuning which performs the LoRA instead of the full fine-tuning in the second stage.
To ensure a similar number of training parameters (i.e., 0.3 \%) with prefix-tuning and LoRA tuning, we train only selected blocks (e.g., only Q-former is trained for the BLIP-2) for the LoRA tuning stage in PT-LoRA.
Other than that, all the training settings are the same as the PT-PEFT.

\section{Additional Experiments}\label{sec:additional_analysis}
\begin{figure*}[t!]
    \centering
    \centering\includegraphics[width=1.0\linewidth]{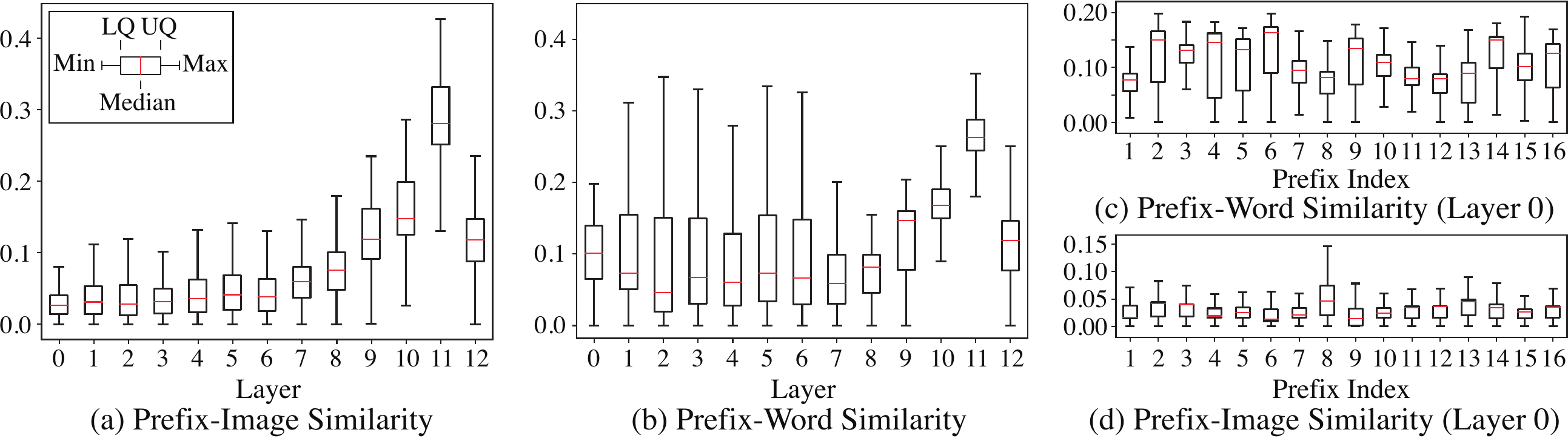}
    \caption{Cosine similarities between prefix-word, and prefix-image feature in image captioning using PT-PEFT.}
    \label{fig:prefix_correlation}
    \vspace{0.4cm}
\end{figure*}

\subsection{Ablation Study}

\paragraph{Prefix Length}
Longer prefixes (i.e., many prefix tokens) involve more trainable parameters, thus assumed to enhance the performance for prefix-tuning~\cite{li2021prefix}.
Figure~\ref{fig:prefix_len_ablation} shows that performance indeed improves as the number of prefix tokens increases, but saturates after a certain point.
Note that previous works on prefix-tuning often used much longer prefix lengths than our PT-PEFT, but since PT-PEFT refines all the parameters, longer prefix seems to be unnecessary for PT-PEFT.

\paragraph{Prefix Encoder}
In order to assess the impact of the prefix encoder design, we conducted ablation studies as summarized in Table~\ref{tab:ablation_prefix_encoder}. 
These experiments were performed on the VQAv2 dataset, following the training step of the PT-PEFT process.
We use the same hyper-parameter settings described in Table~\ref{tab:hyper_param}.
Notably, the results indicate a slight decrease in top-1 accuracy when the prefix type embedding is removed, but there is a significant drop in top-5 accuracy. 
This suggests that the prefix type embedding plays an important role in improving performance.
Furthermore, when the MLP block is removed, top-5 accuracy experiences a considerable decline.
This demonstrates that the prefix encoder contributes to the overall performance of the model, highlighting its importance in capturing and encoding essential information for VQA tasks.

\begin{table*}[t]
\setlength\heavyrulewidth{1.5pt}
\renewcommand{\arraystretch}{1.0}
\centering
\resizebox{\linewidth}{!}{
\begin{tabular}{ccccccccccccc}
\toprule
 & \multicolumn{12}{c}{Alternation Steps} \\
 & \multicolumn{3}{c}{1}  & \multicolumn{3}{c}{2}  & \multicolumn{3}{c}{3}  & \multicolumn{3}{c}{4}  \\
\cmidrule(r){2-4} \cmidrule(r){5-7} \cmidrule(r){8-10} \cmidrule(r){11-13} 
                  & BLEU-4 & CIDEr & SPICE & BLEU-4 & CIDEr & SPICE & BLEU-4 & CIDEr & SPICE & BLEU-4 & CIDEr & SPICE \\
\midrule \midrule
w/ Prefix               & 41.3   & 139.3 & 24.6  & 33.2   & 115.1 & 20.7  & 23.7   & 90.0  & 16.8  & 20.6   & 67.4  & 13.8  \\
\midrule
- Prefix                & 22.9   & 75.0  & 15.3  & 21.5   & 73.8  & 14.9  & 21.2   & 71.3  & 14.5  & 20.6   & 67.4  & 13.8 
\\ \bottomrule
\end{tabular}
}
\caption{Alternation training experiments on COCO image captioning.}
\label{tab:alternation_training}
\end{table*}

\paragraph{Alternation Training}
We conduct experiments to see whether the alternation training can further enhance the performance.
As shown in Table~\ref{tab:alternation_training}, we found that prefix-tuning fails to learn the context necessary for the task during the alternation training.
Even if the initial prefix-tuning is successful (see train alternation step 1), the knowledge learned from the pre-trained model during this phase is lost (see train alternation steps 4, prefix is no longer affecting the output). 
This loss may be attributed to retraining in the collapsed representation space.
Repeated fine-tuning also causes overfitting and performance degradation (see train alternation steps 4 in Table~\ref{tab:alternation_training}).

\begin{figure}[t!]
    \setlength\heavyrulewidth{1.5pt}
    \renewcommand{\arraystretch}{1.0}
    \begin{subfigure}{0.48\textwidth}
    \centering
    \includegraphics[width=1.0\linewidth]{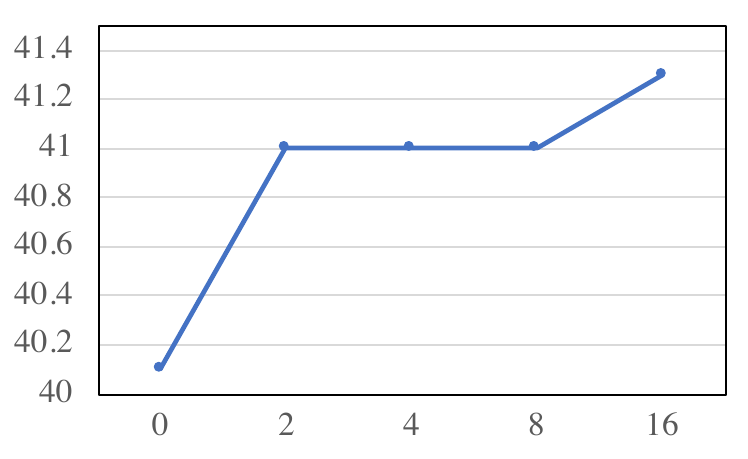}
    \caption{BLEU-4 score}
    \end{subfigure} \vfill
    \begin{subfigure}{0.48\textwidth}
    \centering
    \includegraphics[width=1.0\linewidth]{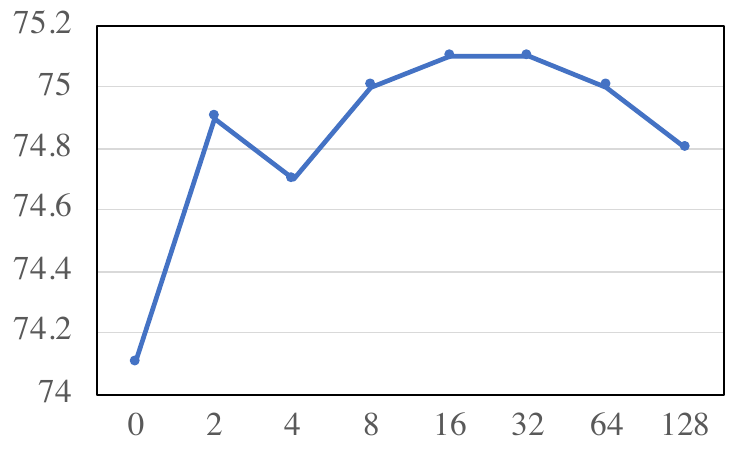}
    \caption{VQAv2 Acc1 (\%)}
    \end{subfigure}
    \caption{Ablation on the prefix length in image captioning and visual question answering. The x-axis indicates the number of prefix tokens used.}
    \label{fig:prefix_len_ablation}
\end{figure}

\subsection{Empirical Analysis}

\paragraph{Mimicking Pre-trained Representations}
To gain insights into the learned representations of the prefix during training, we analyze cosine similarity between prefix tokens and image/caption tokens in the PT-PEFT-tuned model (prefix length of 16).
We observe that the cosine similarities between 16 prefix tokens are very low, all below 0.09.

Furthermore, we find that the correlation between prefix-image and prefix-word increased across the different layers (see Figure~\ref{fig:prefix_correlation}~(a) and~(b)).
Interestingly, the prefix-word similarities (0.1-0.2) are higher than prefix-image similarities (0.0-0.05), especially in lower layers (see Figure~\ref{fig:prefix_correlation}~(c) and~(d)).
This suggests that the prefix maintains its representation space from pre-training by acquiring quasi-orthogonal bases that are relatively closer to pre-trained text features.
However, in higher layers, the prefix-image similarities (0.2-0.4) are higher than prefix-text similarities (0.2-0.35) (see Figure~\ref{fig:prefix_correlation}~(a) and~(b)).
These results clearly indicate that the feature of the image is converted to language space through the interaction with prefix vectors.
\begin{figure*}[t!]
    \centering
    \begin{subfigure}[b]{\textwidth}
    \caption*{\normalsize{VINVL (ResNeXt-152 C4 + BERT-Base)}}
    \vspace{0.2cm}
    \includegraphics[width=1.0\textwidth]{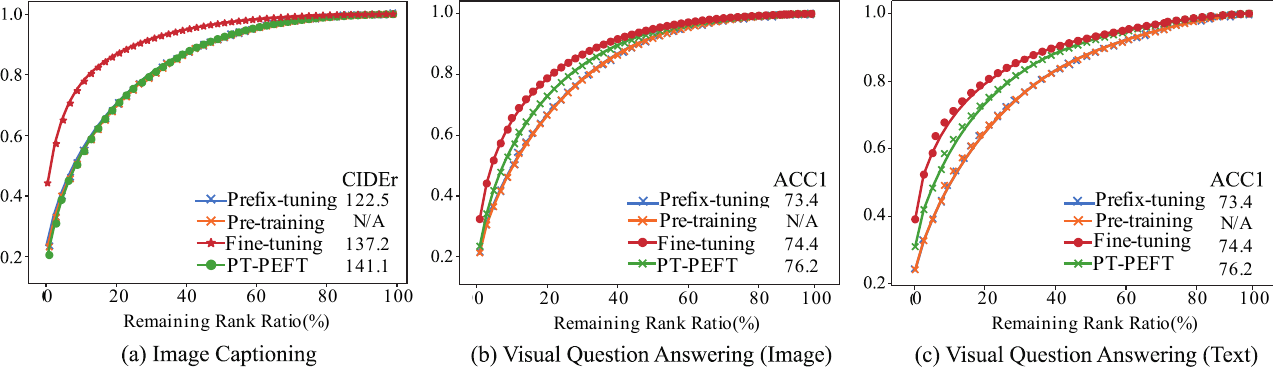}
    \label{fig:svd_ranks_vinvl}
    \end{subfigure}
    \vspace{-0.9cm}
    \caption{Accumulated and normalized singular values of feature vectors extracted from the last layer of VINVL.}
    \label{fig:svd_vinvl}
\end{figure*}
\paragraph{SVD Experiments} We conduct experiments with SVD analysis as in Figure 4 on the VINVL (see Figure~\ref{fig:svd_vinvl}). The results in VINVL also show that representation collapse (i.e., most singular values of the representation matrix are close to zero) in the fine-tuned model while the representation space is preserved (i.e., most singular values are the same) in PT-Full-Finetuning (PT-FT) or PT-LoRA model.

\paragraph{More Qualitative Examples}
Figures~\ref{fig:caption_examples} and~\ref{fig:vqa_examples} show examples of generated captions on the COCO Karpathy test split and VQAv2 valid set, respectively.
We visualize representative images and corresponding captions generated by two models trained using PT-PEFT and fine-tuning.
Compared to the fine-tuned model, the PT-PEFT-tuned model demonstrates a strong ability to capture important details for enriching generated captions.
For example, the proposed method enables extracting proper object-related attributes such as `cut in half', `in the mirror', `in front of', and `a red and yellow'. 
Similarly, in VQA, the predictions from PT-PEFT are more consistent with the answer, and there is a high correlation within the top-5 candidates.
In contrast, the predicted topmost answers after only applying the fine-tuning are much less similar to each other, implying that the learned word representations are lost.
These observations can be attributed to the rank of the feature matrix, as the high-rank features produced by PT-PEFT contain semantically rich information.

\paragraph{Zero-shot Qualitative Example}
To provide a more comprehensive understanding of the qualitative differences between zero-shot, prefix-tuned, and fine-tuned models, we present additional examples in Table~\ref{tab:caption_comparison}. 
These examples illustrate how fine-tuned models, despite achieving high metric scores, may overlook important visual details, resulting in captions that are shorter and more simplified compared to those generated by prefix-tuning and zero-shot approaches.

\begin{table}[t!]
 
    \centering
    \setlength\heavyrulewidth{1.5pt}
    \renewcommand{\arraystretch}{1.2}
    \resizebox{\linewidth}{!}{
    \begin{tabular}{lcccc}
        \toprule
        \multirow{2}{*}{} & \multicolumn{2}{c}{Prefix-tuning Stage} & \multicolumn{2}{c}{Fine-tuning Stage} \\
        \cmidrule(l){2-3} \cmidrule(l){4-5}
         & Acc1 & \multicolumn{1}{c}{Acc5} & Acc1 & \multicolumn{1}{c}{Acc5} \\
        \midrule
        \midrule
        \textbf{PT-PEFT} & \textbf{73.8} & \textbf{93.1} & \textbf{75.2} & \textbf{93.3}\\
        \midrule \midrule
        - Prefix Type Embedding & 73.6 & 90.6 & 74.8 & 91.0\\
        \midrule
        - Prefix MLP & 73.3 & 90.3 & 74.9 & 90.8\\
        \midrule
        - Prefix Encoder & 73.3 & 90.3 & 74.7 & 90.7\\
        \bottomrule
    \end{tabular}
    }
    \caption{Ablation of prefix-encoder implementation on VQAv2 validation split.}
    \label{tab:ablation_prefix_encoder}
    \vspace{-0.2cm}
\end{table}



\section{Discussion}\label{sec:Discussion}
\subsection{Simply Adding Parameters Helps?}
One might assume that the performance enhancement is simply a result of adding additional parameters during fine-tuning. 
However, it is important to note that increasing the number of parameters (i..e, stacking more layers) does not necessarily expand the representation space. 
Intuitively, if we consider a linear transformation where $\mathbf{Y} = \mathbf{W}\mathbf{X}$, with $\mathbf{W}$ as the layer weight and $\mathbf{X}$ as the input, then the rank of $\mathbf{Y}$ is limited by the minimum rank between $\mathbf{W}$ and $\mathbf{X}$ (i.e., $\text{rank}(\mathbf{Y}) \leq \min(\text{rank}(\mathbf{W}), \text{rank}(\mathbf{X}))$).
This means that simply adding more layers would not contribute to avoiding representation collapse.
Moreover, previous research has demonstrated that incorporating more complex layers can lead to a faster collapse in rank \citep{dong2021loserank}.

\subsection{Expressive Power vs. Semantic Richness?}
`Expressive power of parameters' refers to a model’s ability (complexity and size) to adjust its weights to fit a new downstream task. 
On the other hand, a `semantically rich feature representation space' or `high-rank feature' refers to the capability of a model to capture informative features that exhibit strong generalization across different tasks.

To maximize the downstream performance, both `expressive power' and `semantic richness' are important. 
Our experiments show that prefix-tuning, which only tunes a few parameters, has limited expressive power but is good at preserving a semantically rich feature representation space. 
In contrast, fine-tuning, an approach to modify all parameters, has greater expressive power but might distort the representation space, resulting in lower rank and reduced semantic richness compared to a pre-trained model.

Our findings (including SVD analysis and task performance comparison) are consistent with the previous analyses on fine-tuning where `fine-tuning makes the space simpler'~\cite{zhou2021closer} and `simplified space yields lower performance to out-of-domain (OOD) data (bad generalization)'~\cite{kumar2022finetune_distort}. 
In summary, the goal of PT-PEFT is to take advantage of both expressive power and the preservation of semantic richness of the feature representation space.

\subsection{How Prefix-Tuning Preserves the Representation Space?}
To elucidate how prefix-tuning preserves the representation space, we analytically compare the rank of the representation space (i.e., vector space) after applying the attention operation in both fine-tuned and prefix-tuned models.

In a Transformer model, information from the input tokens of the input sequence is mixed exclusively through self-attention. The other components in the Transformer, such as the feed-forward network, are token-wise operators and thus are not affected by prefix tokens. Specifically, for a given input sequence $\text{X} = [\mathbf{x}_0; \dots; \mathbf{x}_N]$, the output of self-attention is the weighted sum of the value matrix $\text{X} \text{W}_V$, where the weights are the attention scores:
\begin{equation}
f(\text{X}) = \sigma(\text{W}_Q \text{X} \text{X}^T \text{W}_K^T) \text{X} \text{W}_V
\end{equation}
where $\sigma$ denotes the softmax function.
In the case of prefix-tuning, the self-attention function is reformulated to incorporate a learnable prefix matrix $\text{P}$:
\begin{equation}
f_{\text{Prefix}}(\text{X}) = \sigma(\text{W}_Q [\text{X}; \text{P}] [\text{X}; \text{P}]^T \text{W}_K^T) \text{X} \text{W}_V
\end{equation}
Here, only the number of input tokens increases while the model parameters remain unchanged.

Considering the rank of the matrix product, which satisfies the inequality $rank(\text{A}\text{B}) \leq \min(rank(\text{A}), rank(\text{B}))$, the rank of the self-attention output is bounded by:
\begin{align}
& rank\left(f(\text{X})\right) \nonumber \\
&\leq \min\left(|\text{X}|, rank\left(\text{X} \text{W}_V\right)\right)
\end{align}
\begin{align}
& rank\left(f_{\text{prefix}} (\text{X})\right) \nonumber \\
&\leq \min( |\text{X}| + |\text{P}|, rank\left(\left[\text{X}; \text{P}\right] \text{W}_V\right))
\end{align}
Assuming the softmax output is full rank, this indicates that the upper bound of the rank is at least as large as the rank of the pre-trained representation space, provided that the parameters remain unchanged:
\begin{align}
&\min(|\text{X}|, rank(\text{X}\text{W}_V)) \nonumber \\ 
&\leq \min(|\text{X}| + |\text{P}|, rank([\text{X}; \text{P}]\text{W}_V))
\end{align}

This analysis suggests that prefix-tuning can maintain or even enhance the semantic richness of the feature representation space by preserving the rank, whereas fine-tuning can reduce the rank, thereby diminishing the semantic richness.

\clearpage
\begin{table*}[ht]
    \centering
    \setlength\heavyrulewidth{1.5pt}
    \resizebox{0.9\textwidth}{!}{
    \begin{tabular}{>{\raggedright}p{2cm} >{\raggedright}p{4.5cm} >{\raggedright}p{4.5cm} >{\raggedright\arraybackslash}p{4.5cm}}
        \toprule
        \textbf{COCO Image ID} & \textbf{Zero-Shot} & \textbf{Finetune} & \textbf{Prompt} \\
        \midrule
        272117 & ``a group of people sitting around a table with a birthday cake in front of them'' & ``a group of people sitting around a table with a cake'' & ``a group of people sitting around a table with a birthday cake in front of them'' \\
        503392 & ``two horses in an arena with a person riding on the back of one of the horses'' & ``two horses in an arena with a person riding one of the horses'' & ``two horses in an arena with a person riding on the back of one of the horses'' \\
        60467 & ``a lunch tray with a breakfast sandwich, orange juice, and a glass of milk'' & ``a lunch tray with a sandwich, orange juice, and a glass of milk'' & ``a tray of food on a table'' \\
        544471 & ``a man and a woman sitting on a brick wall with a laptop in front of them'' & ``a woman and a boy sitting on steps with a laptop'' & ``a man and a woman posing with a laptop'' \\
        117170 & ``two pizza rolls sitting on a counter with a sign that says `pizza rolls' '' & ``two pizza rolls sitting on top of a silver platter'' & ``two pizza rolls on a silver platter with a sign that says `pizza rolls' '' \\
        235644 & ``a group of people working on a person on a stretcher at a train station'' & ``a group of people on a platform next to a train'' & ``three people helping a person on a stretcher on a train platform'' \\
        514607 & ``an umbrella on a beach with rocks and a body of water in the background'' & ``an umbrella on a rocky beach with the ocean in the background'' & ``a beach with a beach umbrella in the foreground and the ocean in the background'' \\
        89541 & ``a container of food with strawberries, blueberries, and a muffin in it'' & ``a bowl filled with fruit and muffins on a table'' & ``a yellow container with strawberries, blueberries, and a muffin in it'' \\
        477470 & ``a street at night with traffic lights and a building in the background'' & ``a traffic light on a city street at night'' & ``a street at night with traffic lights and a building in the background.'' \\
        529004 & ``a car driving down a road with a herd of cows on the side of the road'' & ``a herd of cattle crossing a road in front of a car'' & ``a car driving down a road with a herd of cows on the side of the road'' \\
        545407 & ``an airplane flying in the sky with a clear blue sky in the background'' & ``an airplane flying through a clear blue sky'' & ``an airplane flying in the sky with a blue sky behind it'' \\
        255036 & ``an intersection with traffic lights and a building in the background'' & ``a traffic light sitting on the corner of a street'' & ``a traffic light at an intersection with a building in the background'' \\
        276146 & ``a pizza on a cutting board with a glass of wine and a bottle of wine'' & ``a pizza sitting on a cutting board next to a bottle of wine'' & ``a pizza on a cutting board with a glass of wine next to it'' \\
        62554 & ``some food on a table with a bowl of broccoli and a bowl of asparagus'' & ``a table topped with bowls of food and plates of food'' & ``a bowl of broccoli and a bowl of asparagus on a table'' \\
        554980 & ``a red school lunch tray with a sandwich, orange, and a glass of milk'' & ``a red plastic tray with a sandwich, fruit, and a glass of milk'' & ``a red tray with food on it'' \\
        290951 & ``people walking in a building with umbrellas hanging from the ceiling'' & ``people walking under colorful umbrellas in a building'' & ``umbrellas suspended from the ceiling of a building'' \\
        299039 & ``a plate of food on a table with a vase of flowers in the background'' & ``a plate of food on a table with a vase of flowers'' & ``a plate of food on a table with a vase of flowers in the background'' \\
        379842 & ``a wii game with a wii remote and nintendo super mario galaxy 2 game'' & ``a wii game and controller sitting on a table'' & ``a wii remote and nintendo super mario galaxy 2 game'' \\
        \bottomrule
    \end{tabular}}
    \caption{Comparison of Captioning Methods on COCO Dataset}
    \label{tab:caption_comparison}
\end{table*}

\newpage
\setlength{\tabcolsep}{15pt}
\begin{table*}[t]
\centering
\setlength\heavyrulewidth{1.5pt}
\resizebox{\linewidth}{!}{
\begin{tabular}{lccccc}
\toprule
Training method & \multicolumn{1}{c}{Total train epoch} & \multicolumn{1}{c}{Warmup epoch} & \multicolumn{1}{c}{Max LR}   & \multicolumn{1}{c}{Batch size} & \multicolumn{1}{c}{Weight decay} \\
\midrule
\midrule
\multicolumn{6}{c}{COCO IC BASE}                                                                                                                                                             \\

Prefix-tuning  & 30                                    & 3                                & 1.00E-05                     & 1024                           & 0.2                              \\
CE             & 40                                    & 12                               & 1.00E-05                     & 1024                           & 0.2                              \\
SCST           & 75                                    & 15                               & 3.00E-06                     & 128                            & 0.2                              \\
\midrule
\multicolumn{6}{c}{COCO IC LARGE}                                                                                                                                                            \\
Prefix-tuning             & 30                                    & 3                                & 1.00E-05                     & 512                            & 0.2                              \\
CE             & 30                                    & 6                                & 3.00E-06                     & 512                            & 0.2                              \\
SCST           & 50                                    & 10                               & 3.00E-06                     & 192                            & 0.1                              \\
\midrule \midrule
\multicolumn{6}{c}{Flickr30k IC BASE}                                                                                                                                                        \\
Prefix-tuning             & \multicolumn{1}{c}{30}                & \multicolumn{1}{c}{0}            & \multicolumn{1}{c}{5.00E-05} & \multicolumn{1}{c}{512}        & \multicolumn{1}{c}{0.1}          \\
Fine-tuning             & \multicolumn{1}{c}{70}                & \multicolumn{1}{c}{0}            & \multicolumn{1}{c}{1.00E-05} & \multicolumn{1}{c}{512}        & \multicolumn{1}{c}{0.15}         \\
\midrule
\multicolumn{6}{c}{Flickr30k IC LARGE}                                                                                                                                                       \\
Prefix-tuning             & \multicolumn{1}{c}{30}                & \multicolumn{1}{c}{0}            & \multicolumn{1}{c}{5.00E-05} & \multicolumn{1}{c}{512}        & \multicolumn{1}{c}{0.1}          \\
Fine-tuning             & \multicolumn{1}{c}{70}                & \multicolumn{1}{c}{0}            & \multicolumn{1}{c}{3.00E-05} & \multicolumn{1}{c}{512}        & \multicolumn{1}{c}{0.15}         \\
\midrule \midrule
\multicolumn{6}{c}{VQA BASE}                                                                                                                                                                 \\
Prefix-tuning  & 50                                    & 0                                & 1.00E-04                     & 512                            & 0.05                             \\
Fine-tuning    & 25                                    & 3                                & 1.00E-05                     & 512                            & 0.05                             \\
\midrule
\multicolumn{6}{c}{VQA LARGE}                                                                                                                                                                \\
Prefix-tuning  & 50                                    & 0                                & 5.00E-05                     & 512                            & 0.05                             \\
Fine-tuning    & 25                                    & 3                                & 5.00E-06                     & 512                            & 0.05                             \\
\midrule \midrule
\multicolumn{6}{c}{GQA BASE}                                                                                                                                                                 \\
Prefix-tuning  & 5                                     & 0.5                              & 1.00E-04                     & 512                            & 0.05                             \\
Fine-tuning    & 5                                     & 0.5                              & 1.00E-05                     & 512                            & 0.05                    \\    
\bottomrule
\end{tabular}}
\caption{Training hyper-parameters for VINVL. PT-PEFT is trained with the same hyper-parameter with Fine-tuning (CE) in the table. Image size of 640x480 is used.}
\label{tab:hyper_param}
\end{table*}
\setlength{\tabcolsep}{6pt}

\newpage
\setlength{\tabcolsep}{12pt}
\begin{table*}[t]

\centering
\setlength\heavyrulewidth{1.5pt}
\resizebox{\linewidth}{!}{
\begin{tabular}{lccccc}
\toprule
Training method & \multicolumn{1}{c}{Total train epoch} & \multicolumn{1}{c}{Warmup epoch} & \multicolumn{1}{c}{Max LR}   & \multicolumn{1}{c}{Batch size} & \multicolumn{1}{c}{Weight decay} \\
\midrule
\midrule
\multicolumn{6}{c}{COCO IC}                                                                                                                                                            \\
Prefix-tuning             & 10                                    & 0                                & 1.00E-03                     & 16                            & 0.01                              \\
LoRA & 5                                    & 0                                & 1.00E-03                     & 16                            & 0.01                              \\
Fine-tuning             & 5                                    & 0                                & 1.00E-03                     & 16                            & 0.15                              \\
PT-FT (2nd Stage)            & \multicolumn{1}{c}{10}                & \multicolumn{1}{c}{0}            & \multicolumn{1}{c}{1.00E-05} & \multicolumn{1}{c}{16}        & \multicolumn{1}{c}{0.15}         \\
PT-LoRA (2nd Stage)             & \multicolumn{1}{c}{10}                & \multicolumn{1}{c}{0}            & \multicolumn{1}{c}{1.00E-05} & \multicolumn{1}{c}{16}        & \multicolumn{1}{c}{0.15}         \\
\midrule \midrule

\multicolumn{6}{c}{Flickr30k IC}                                                                                                                                                       \\
Prefix-tuning             & \multicolumn{1}{c}{5}                & \multicolumn{1}{c}{0}            & \multicolumn{1}{c}{1.00E-03} & \multicolumn{1}{c}{16}        & \multicolumn{1}{c}{0.01}          \\
LoRA             & \multicolumn{1}{c}{5}                & \multicolumn{1}{c}{0}            & \multicolumn{1}{c}{1.00E-03} & \multicolumn{1}{c}{16}        & \multicolumn{1}{c}{0.01}          \\
Fine-tuning             & \multicolumn{1}{c}{5}                & \multicolumn{1}{c}{0}            & \multicolumn{1}{c}{1.00E-03} & \multicolumn{1}{c}{16}        & \multicolumn{1}{c}{0.15}         \\
PT-FT (2nd Stage)             & \multicolumn{1}{c}{10}                & \multicolumn{1}{c}{0}            & \multicolumn{1}{c}{1.00E-05} & \multicolumn{1}{c}{16}        & \multicolumn{1}{c}{0.15}         \\
PT-LoRA (2nd Stage)              & \multicolumn{1}{c}{10}                & \multicolumn{1}{c}{0}            & \multicolumn{1}{c}{1.00E-05} & \multicolumn{1}{c}{16}        & \multicolumn{1}{c}{0.15}         \\
\midrule \midrule
\multicolumn{6}{c}{VQA} \\

Prefix-tuning  & 50                                    & 0                                & 1.00E-04                     & 512                            & 0.05                             \\
LoRA  & 50                                    & 0                                & 1.00E-04                     & 512                            & 0.05                             \\
Fine-tuning   & 25                                    & 3                                & 1.00E-05                     & 512                            & 0.05                             \\
PT-FT (2nd Stage)             & \multicolumn{1}{c}{10}                & \multicolumn{1}{c}{0}            & \multicolumn{1}{c}{1.00E-05} & \multicolumn{1}{c}{16}        & \multicolumn{1}{c}{0.15}         \\
PT-LoRA (2nd Stage)              & \multicolumn{1}{c}{10}                & \multicolumn{1}{c}{0}            & \multicolumn{1}{c}{1.00E-05} & \multicolumn{1}{c}{16}        & \multicolumn{1}{c}{0.15}         \\
\bottomrule
\end{tabular}}
\caption{Training hyper-parameters for OFA.  Image size of 480x480 is used.}
\label{tab:hyper_param}
\end{table*}
\setlength{\tabcolsep}{6pt}

\newpage
\setlength{\tabcolsep}{13pt}
\begin{table*}[t]

\centering
\setlength\heavyrulewidth{1.5pt}
\resizebox{\linewidth}{!}{
\begin{tabular}{lccccc}
\toprule
Training method & \multicolumn{1}{c}{Total train epoch} & \multicolumn{1}{c}{Warmup Steps} & \multicolumn{1}{c}{Max LR}   & \multicolumn{1}{c}{Batch size} & \multicolumn{1}{c}{Weight decay} \\
\midrule
\midrule
\multicolumn{6}{c}{COCO IC} \\
Prefix-tuning & 5  & 5000 & 5.00E-05 & 128 & 0.05 \\
LoRA & 5 & 5000 & 1.00E-04 & 128 & 0.05 \\
Fine-tuning & 5 & 5000 & 1.00E-05 & 128 & 0.05 \\
\midrule \midrule
\multicolumn{6}{c}{Flickr30k IC} \\
Prefix-tuning & 5 & 5000 & 5.00E-05 & 128 & 0.05 \\
LoRA & 5 & 5000 & 1.00E-04 & 128 & 0.05 \\
Fine-tuning & 5 & 5000 & 1.00E-05 & 128 & 0.05 \\
\midrule \midrule
\multicolumn{6}{c}{VQA} \\
Prefix-tuning  & 5 & 0 & 5.00E-05 & 512 & 0.05 \\
LoRA  & 5 & 0 & 1.00E-04 & 128 & 0.05 \\
Fine-tuning   & 5 & 0 & 1.00E-03 & 128 & 0.05 \\
\bottomrule
\end{tabular}}
\caption{Training hyper-parameters for BLIP-2. PT-PEFT and PT-LoRA are trained with the same hyper-parameter with LoRA and Fine-tuning in the table. Image size of 224x224 is used.}
\label{tab:hyper_param_blip}
\end{table*}
\setlength{\tabcolsep}{6pt}
\begin{figure*}[t!]
     \centering
     \begin{subfigure}[b]{0.40\textwidth}
         \centering
         \includegraphics[width=\textwidth]{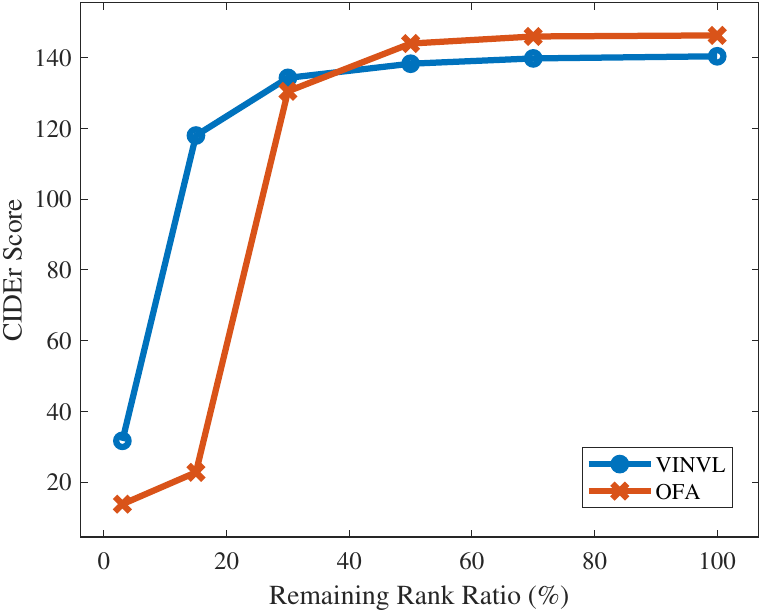}
         \caption{CIDEr scores on Image Captioning}
         \label{fig:rank_reduction_graph}
     \end{subfigure}
     \hfill
     \begin{subfigure}[b]{0.55\textwidth}
         \centering
         \includegraphics[width=\textwidth]{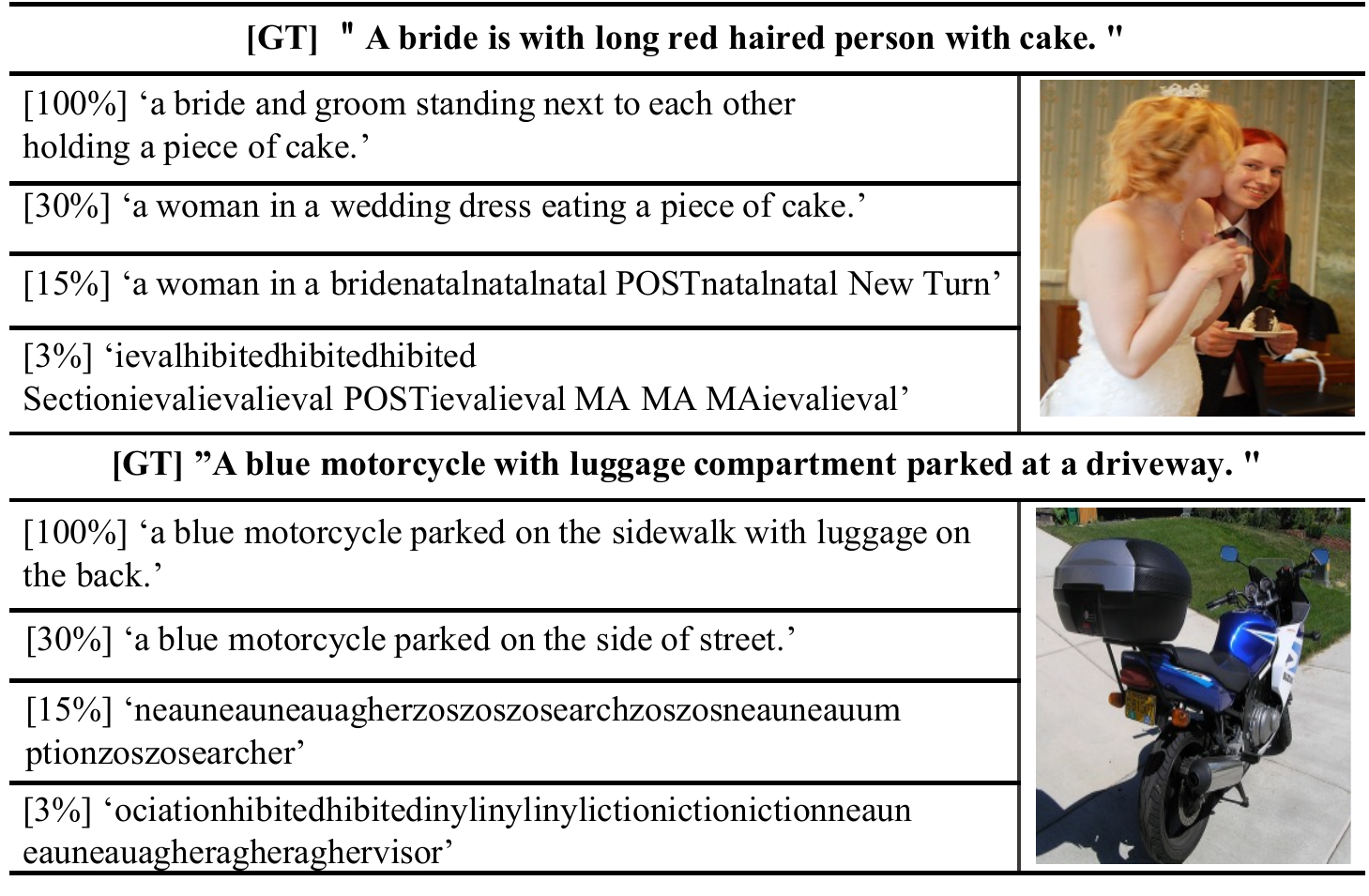}
         \caption{Generated caption examples}
         \label{fig:rank_reduction_examples}
     \end{subfigure}
    \caption{The effect of rank reduction on COCO image captioning performance. The percentage in (b) denotes the remaining rank ratio.}
\label{fig:three graphs}
\vspace{-0.4cm}
\end{figure*}
\newpage
\begin{figure*}[p]
\centering
\includegraphics[width=0.9\linewidth]{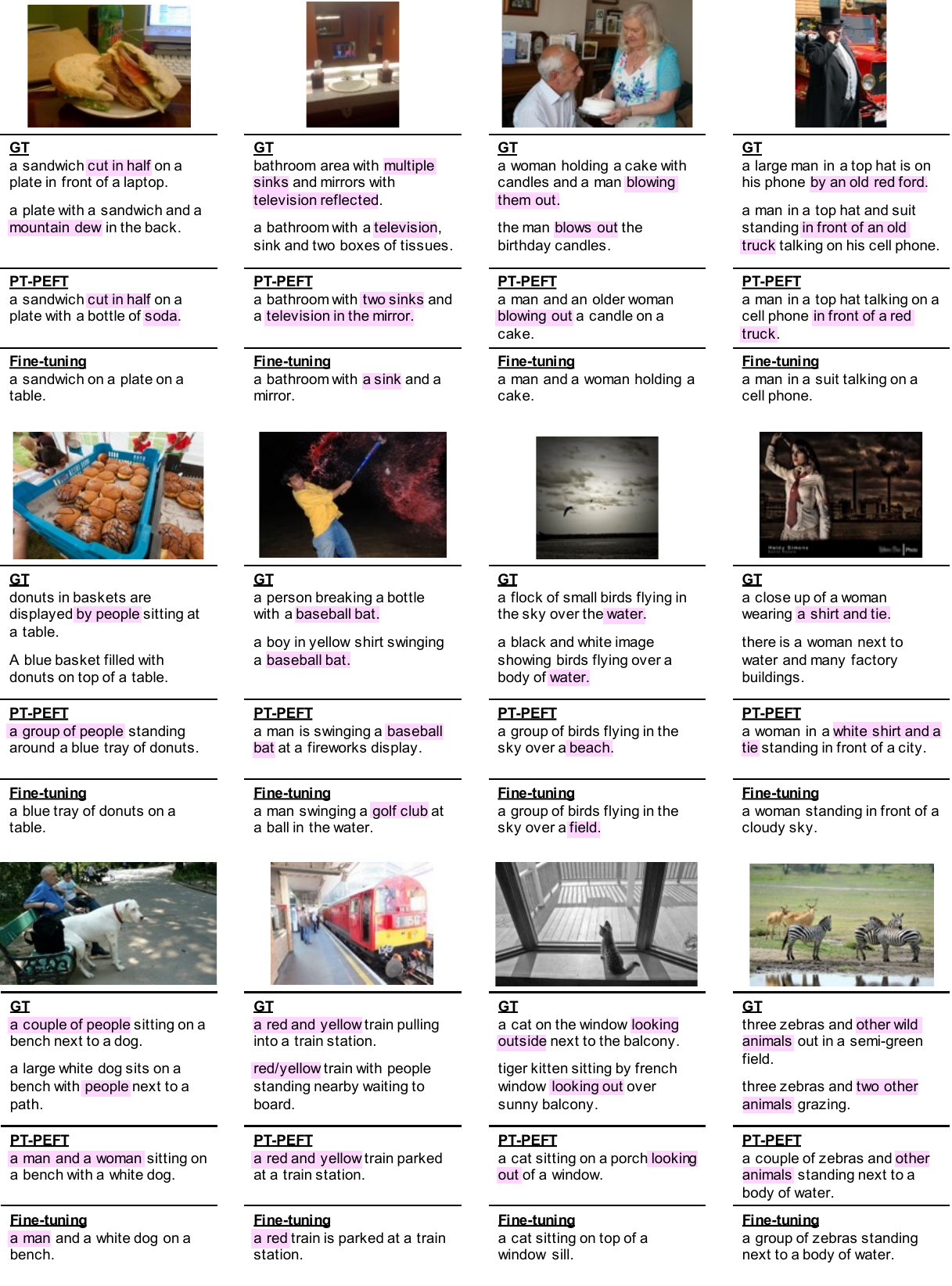}
\caption{Qualitative examples of generated captions on COCO Karpathy test split. 
\textbf{GT}: the ground-truth captions.
}
\label{fig:caption_examples}
\end{figure*}
\newpage
\begin{figure*}[p]
\centering
\includegraphics[width=0.9\linewidth]{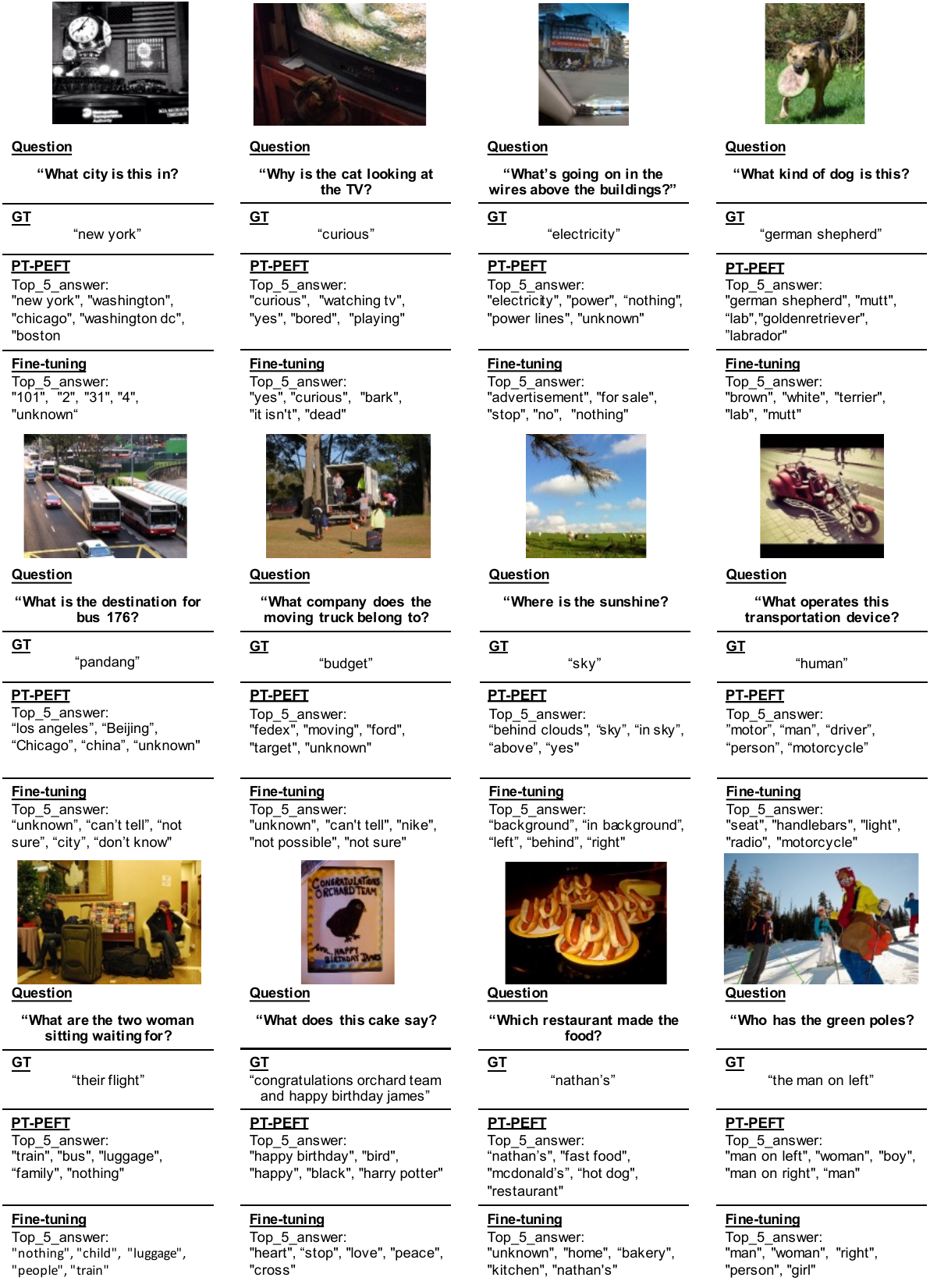}
\caption{Qualitative examples of generated captions on VQAv2 validation split. 
\textbf{GT}: the ground-truth answer.
}
\label{fig:vqa_examples}
\end{figure*}

\end{document}